\documentclass[10pt,twocolumn,letterpaper]{article}

\usepackage[pagenumbers]{iccv} 

\definecolor{iccvblue}{rgb}{0.21,0.49,0.74}
\usepackage[pagebackref,breaklinks,colorlinks,allcolors=iccvblue]{hyperref}

\usepackage{multicol}
\usepackage{multirow}
\usepackage{bm}

\usepackage{algorithm}      
\usepackage{algorithmic}
\usepackage{amsmath}        
\usepackage{amssymb}        
\usepackage{amsfonts}       

\usepackage{xcolor}      
\usepackage{listings}    

\def\paperID{1064} 
\def\confName{ICCV}
\def\confYear{2025}

\title{A Diffusion-Based Framework for Terrain-Aware Remote Sensing Image Reconstruction}

\author{
    Zhenyu Yu\\
    Universiti Malaya\\
    {\tt\small yuzhenyuyxl@foxmail.com}
    \and
    Mohd Yamani Inda Idris\\
    Universiti Malaya\\
    {\tt\small yamani@um.edu.my}
    \and
    Pei Wang$^*$\\
    Kunming University of \\Science and Technology\\
    {\tt\small peiwang@kust.edu.cn}
}

\begin{document}

\twocolumn[{
\renewcommand\twocolumn[1][]{#1}
\maketitle
\begin{center}
    \captionsetup{type=figure}
    \includegraphics[width=1.0\linewidth]{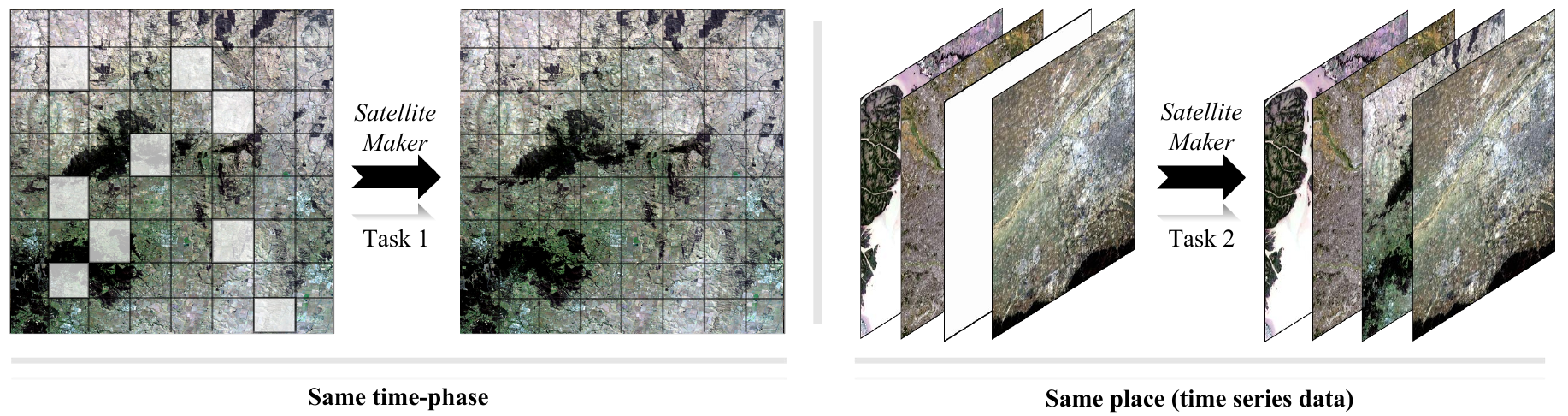} 
    \caption{Motivation. We proposed \textit{SatelliteMaker} as a method utilizing controlled diffusion models to address critical challenges in remote sensing imagery, including cloud occlusion, missing data, and distribution discripancy.} 
    \label{fig_problem}
\end{center}
}]

\begin{abstract}

Remote sensing imagery is essential for environmental monitoring, agricultural management, and disaster response. However, data loss due to cloud cover, sensor failures, or incomplete acquisition—especially in high-resolution and high-frequency tasks—severely limits satellite imagery's effectiveness. Traditional interpolation methods struggle with large missing areas and complex structures. Remote sensing imagery consists of multiple bands, each with distinct meanings, and ensuring consistency across bands is critical to avoid anomalies in the combined images.  
This paper proposes \textit{SatelliteMaker}, a diffusion-based method that reconstructs missing data across varying levels of data loss while maintaining spatial, spectral, and temporal consistency. We also propose Digital Elevation Model (DEM) as a conditioning input and use tailored prompts to generate realistic images, making diffusion models applicable to quantitative remote sensing tasks. Additionally, we propose a VGG-Adapter module based on Distribution Loss, which reduces distribution discrepancy and ensures style consistency. Extensive experiments show that \textit{SatelliteMaker} achieves state-of-the-art performance across multiple tasks. The dataset and code will be released upon acceptance.

\end{abstract}    
\section{Introduction}

Remote sensing imagery plays a critical role in various applications such as environmental monitoring, agricultural management, urban planning, and disaster response \cite{zhu2017deep, zhang2010multi,yu2024capan}. However, these images often suffer from missing data due to cloud cover, sensor malfunctions, or incomplete data acquisition, which significantly hinders the analysis and interpretation of satellite imagery, particularly for high-resolution or high-temporal-frequency tasks \cite{belgiu2016random, gorelick2017google,tan2023spatiotemporal,yu2023impact}. Traditional methods, such as interpolation, struggle to effectively handle large missing areas or complex image structures \cite{li2023satellite,yang2020computer,wang2024multi}.

\textbf{Challenges.} Accurately reconstructing incomplete remote sensing images is essential for fully utilizing satellite data \cite{li2024deep,yu2025yuan}. Although deep learning approaches, such as Convolutional Neural Networks (CNNs) and Generative Networks, have shown promise in image inpainting \cite{zhang2023fully,yu2025qrs}, they still face challenges in handling large missing areas and maintaining the spatial and spectral consistency of the original imagery \cite{gui2024remote,yu2025guideline}. Additionally, these methods often require large amounts of labeled data for training and may introduce artifacts in the generated images \cite{wang2023score,yu2025ai}.

\textbf{Diffusion models}, which generate high-quality images by progressively denoising data, have demonstrated impressive results in natural image reconstruction. However, their application in remote sensing remains underexplored \cite{ho2020denoising, lugmayr2022repaint}. Leveraging the strengths of diffusion models offers an opportunity to overcome the limitations of existing methods and provide more effective solutions for satellite image completion \cite{xiao2023ediffsr}.

\textbf{Motivation}. This research enhances the reliability and usability of satellite imagery by addressing incomplete data. Accurate reconstruction improves data quality for environmental monitoring, urban planning, and long-term studies like climate change analysis. We propose \textbf{\textit{SatelliteMaker}}, a diffusion-based model, tackling both \textbf{spatial gaps} within a fixed period and \textbf{temporal gaps} at specific time points, ensuring spatial-temporal consistency (Figure \ref{fig_problem}).

The key \textbf{contributions} of this work are as follows:
\begin{itemize}
    \item \textbf{SatelliteMaker}: A diffusion-based model for filling missing data in remote sensing images, demonstrating stable performance for missing ratios (10\% to 50\%).
    \item \textbf{DEM-guided generation}: The model incorporates Digital Elevation Models (DEMs) as conditional inputs and designs tailored prompts to control and guide content generation, enabling the application of diffusion models to quantitative remote sensing tasks.
    \item \textbf{VGG-adapter module}: A distribution loss-based module that effectively reduces data distribution discrepancies and ensures color consistency in the generated images, achieving state-of-the-art results in two tasks.
\end{itemize}

\begin{figure*}[!ht]
    \centering
    \includegraphics[width=1.0\linewidth]{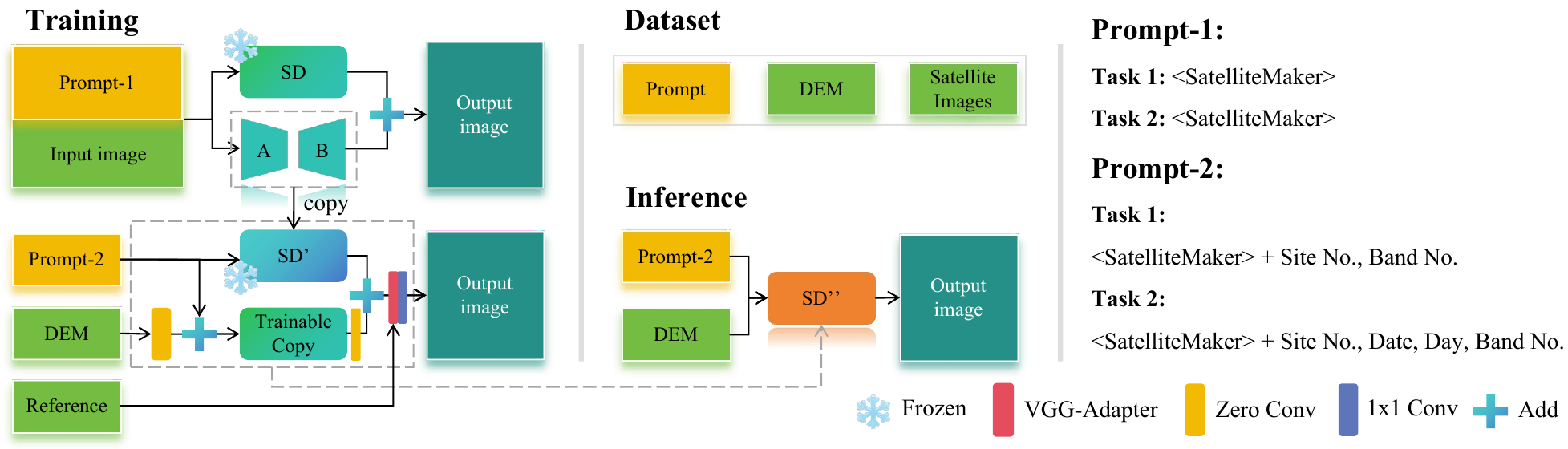}
    \caption{The model's training and inference pipeline consists of two main stages. We design the training phase to develop a diffusion model specifically for remote sensing imagery. A guidance mechanism is incorporated to control the generation process, ensuring stable and accurate image outputs, which are essential for subsequent quantitative remote sensing analysis. In the inference phase, we employ an adaptation module to ensure that reconstructed regions align with the target domain.}
    \label{fig_architecture}
\end{figure*}

\section{Related Work}

\subsection{Data Issues and Traditional Approaches}

Remote sensing imagery is widely used in earth observation and environmental monitoring, yet data issues such as cloud cover, sensor failures, and temporal discontinuities frequently lead to missing or corrupted data \cite{ju2008availability, xie2016recover, daras2024ambient}. These gaps are particularly problematic for high-resolution imagery and long-term monitoring tasks. 

Traditional methods for filling missing data rely on interpolation techniques such as nearest-neighbor \cite{rukundo2012nearest,xing2022benefit}, bilinear \cite{gribbon2004novel,yan2021implementing}, spline \cite{mckinley1998cubic,sun2023cubic}, and kriging \cite{oliver1990kriging,jang2024kriging} interpolation. While effective for small and smooth missing regions, these methods fail to reconstruct large gaps or preserve complex spatial patterns, often introducing blurriness or spectral inconsistencies \cite{kakar2011exposing,de2024improving}. These challenges have motivated the adoption of machine learning and deep learning methods for more sophisticated image reconstruction.

\subsection{Machine Learning Methods}

In recent years, machine learning and deep learning have revolutionized remote sensing image reconstruction. Convolutional Neural Networks (CNNs) \cite{gu2018recent} have been widely applied in image restoration due to their ability to capture spatial features. Partial Convolutions \cite{liu2018image} and Generative Adversarial Networks (GANs) \cite{goodfellow2014generative} further advanced image completion, with methods such as Contextual Attention GAN \cite{yu2018generative} and the Globally and Locally Consistent Image Completion model \cite{iizuka2017globally} demonstrating strong performance in generating visually plausible image completions.
Variational Autoencoders (VAEs) \cite{kingma2013auto} have also been applied to image restoration by learning latent distributions, though they tend to generate over-smoothed results, limiting their applicability in high-resolution remote sensing imagery \cite{kingma2019introduction}. 

Spatio-temporal models, such as Spatio-Temporal Convolutional Neural Networks (ST-CNN) \cite{he2019stcnn} and Spatio-Temporal Graph Convolutional Networks (ST-GCN) \cite{song2020spatial}, have been introduced to address dependencies in time-series satellite data. However, they struggle to maintain spectral fidelity and often require large annotated datasets, which are scarce in remote sensing.

\subsection{Diffusion Models}

Recent advances in generative modeling, particularly diffusion models, have significantly improved image synthesis tasks \cite{song2020improved,dhariwal2021diffusion,rombach2022high}. These models iteratively add and remove noise to generate high-quality images, with Denoising Diffusion Probabilistic Models (DDPM) \cite{ho2020denoising} demonstrating remarkable success in natural image inpainting.

However, diffusion models face challenges in complex structured data. While methods such as Stable Diffusion \cite{rombach2022high} optimize high-resolution image generation, they do not explicitly address spectral consistency across multiple bands, a critical factor in remote sensing. Additionally, ensuring that generated content aligns with physical ground truth data remains an open problem.

To introduce explicit control, ControlNet \cite{zhang2023adding} integrates external conditioning (e.g., edge maps, depth maps) into diffusion models, enabling stable and structured image generation. GeoSynth \cite{sastry2024geosynth} attempted to use basemaps as conditioning information, but such methods remain limited to qualitative reconstruction rather than precise quantitative remote sensing tasks.

Unlike natural image datasets, remote sensing images involve multi-band spectral information and strict spatial-temporal consistency, which traditional diffusion models do not explicitly handle. While existing generative models focus on visual plausibility, remote sensing imagery demands accurate reconstructions that align with geospatial and spectral properties.

\section{Proposed Method - \textit{SatelliteMaker}}




The overall methodology is illustrated in Figure \ref{fig_architecture}. This study focuses on two main tasks:

\textbf{Task 1:} Data from different geographic locations at the same time point. Specifically, we select 10 diverse locations and use Google Earth Engine (GEE) to download data, utilizing Landsat-8 imagery for comprehensive coverage.

\textbf{Task 2:} Data from different time points at the same location. For this task, we use the EarthNet2021 dataset, which provides a temporal series for a consistent location, facilitating analysis of changes over time.

The process involves following steps. Initially, we train a model using Low-Rank Adaptation (LoRA) by injecting a custom keyword \texttt{\textless SatelliteMaker\textgreater}, enabling the model to focus on specific satellite imagery tasks. Then, ControlNet is employed to further train the model, conditioning it on Digital Elevation Model (DEM) data to guide Stable Diffusion (SD) for enhanced spatial accuracy. Finally, we apply the VGG-Adapter module to align feature distributions and minimize color discrepancies, ensuring consistency in output imagery.

\subsection{Lightweight Model Fine-Tuning}

Fine-tuning large pre-trained models can be computationally expensive and resource-intensive, especially when full model retraining is required. To address this, Low-Rank Adaptation (LoRA) \cite{hu2021lora} allows efficient adaptation by introducing low-rank matrices, significantly reducing the number of trainable parameters. This makes it ideal for resource-constrained scenarios while maintaining model performance. LoRA introduces two smaller matrices, $A \in \mathbb{R}^{d \times r}$ and $B \in \mathbb{R}^{r \times k}$, where $r \ll \min(d, k)$, to update the original weight matrix $W$. The new weight matrix $W'$ is computed as:
\begin{equation}
W' = W + AB
\end{equation}
This approach allows the original weights $W$ to remain unchanged, preserving the pre-trained knowledge while adapting the model to specific tasks with fewer resources. 



\begin{algorithm}[H]
\caption{Training Process}
\label{alg:training_process}
\begin{algorithmic}[1]
\REQUIRE Input data $X$, DEM data $D$, Prompt $P$, Learning rate $\eta$, Epochs $E$
\ENSURE Model parameters $\theta$

\STATE Initialize $\theta$

\FOR{$epoch = 1$ to $E$}
    \FOR{each $(x, d, p) \in (X, D, P)$}
        \STATE Set $c \gets d$ \COMMENT{DEM as condition}
        \STATE Generate $\hat{x} \gets f_{gen}(x, c, p, \theta)$ 
        \STATE Compute loss:
        \STATE \quad $\mathcal{L}_{recon} \gets \frac{1}{N} \sum\limits_{i=1}^{N} || x_i - \hat{x}_i ||_2^2$
        \STATE \quad $\mathcal{L}_{dis} \gets \left\| \frac{1}{N} \sum\limits_{i=1}^{N} \phi(x_i) - \frac{1}{N} \sum\limits_{j=1}^{N} \phi(\hat{x}_j) \right\|^2_\mathcal{H}$
        \STATE \quad $\mathcal{L}_{style} \gets \sum\limits_{l=1}^{L} \frac{1}{4C_l^2 H_l^2 W_l^2} \left\| G^l_{gen} - G^l_{ref} \right\|_F^2$
        \STATE \quad $\mathcal{L}_{total} \gets \mathcal{L}_{recon} + \mathcal{L}_{dis} + \lambda \mathcal{L}_{style}$
        \STATE Update parameters:
        \STATE \quad $\theta \gets \theta - \eta \nabla_{\theta} \mathcal{L}_{total}$
    \ENDFOR
\ENDFOR

\STATE \textbf{Return} $\theta$
\end{algorithmic}
\end{algorithm}

\subsection{Conditional Guidance using DEM Data}

A Digital Elevation Model (DEM) is a digital representation of terrain, describing the earth's surface elevations in raster format, with each pixel representing the elevation at that point \cite{polidori2020digital}. DEMs are widely used in terrain analysis, flood modeling, and soil erosion assessment due to their high accuracy and detailed data layers. Since DEM is closely linked to surface features like slope and terrain morphology, it plays a crucial role in geographic data processing and remote sensing analysis \cite{abrams2020aster}. We use DEM as the conditional input because it enhances spatial consistency and geographic accuracy by providing topographic information that complements remote sensing images. This allows the model to better replicate surface features, especially in areas with complex terrain, and generate more contextually accurate outputs for specific geographic regions.


Let $x$ represent the input data (i.e., satellite imagery), and $c$ represent the conditioning information derived from DEM data. The generative process is described as follows:
\begin{equation} 
p(x|c) = \prod_{t=1}^{T} p(x_t|x_{t-1}, c) \end{equation}
where $x_t$ represents the output generated at each time step, conditioned on the previous output $x_{t-1}$ and the conditioning information $c$. 



To incorporate the conditioning information into the prompt, we modify the input prompt to include spatial and temporal details as follows:\\
\texttt{Task 1: \textless satelliteMaker\textgreater \ Site, Band.}\\
\texttt{Task 2: \textless satelliteMaker\textgreater \ Location, Date, Day, Band.}

The training objective for conditional generation is to minimize the difference between the generated output and the real data, which is achieved using reconstruction loss:
\begin{equation} 
\mathcal{L}_{recon} = \frac{1}{N} \sum_{i=1}^{N} || x_i - \hat{x}_i ||_2^2 
\end{equation}
$N$ is the sample size. By minimizing $\mathcal{L}_{recon}$, we ensure that the generated output closely matches the expected satellite imagery.

This DEM-based conditional guidance approach not only effectively utilizes topographic information but also generates satellite images that are consistent with surface features, enhancing the accuracy and fidelity of generated outputs in complex terrain conditions.

\subsection{Reducing Distribution Shift}

The VGG-Adapter module (Figure \ref{fig_vgg19}) aims to reduce distribution shift and color discrepancies in generated images by preserving the style features of the original image. By incorporating Gram matrices into the VGG-19, this module aligns the style features of the generated image with those of the reference image, ensuring consistency in color and texture. 

VGG-19 consists of 19 layers, including 16 convolutional layers and 3 fully connected layers. In style transfer, we use $conv1_1$, $conv2_1$, $conv3_1$, $conv4_1$, and $conv5_1$ for style representation, and $conv4_2$ for content representation. The weights are set as ${1.0,0.8,0.5,0.3,0.1}$. These varying weights are used to give more emphasis to lower layers, which capture finer texture details, while higher layers capture broader style patterns, contributing less as we move deeper into the network.

\textbf{Distribution Loss.} The distribution loss measures the discrepancy between the distributions of the generated images and the reference images. Let $x_i$ be the feature maps from the reference images and $\hat{x}_i$ be the set of feature maps from the generated images. $N$ is the sample size. The distribution loss is formulated as:
\begin{equation} 
\mathcal{L}_{dis} = \left|\left| \frac{1}{N} \sum_{i=1}^{N} \phi(x_i) - \frac{1}{N} \sum_{j=1}^{N} \phi(\hat{x}_j) \right|\right|^2_\mathcal{H}
\end{equation}
where $\phi$ is a feature map to a reproducing kernel Hilbert space (RKHS), such as the Gaussian kernel. This loss minimizes the discrepancy between generated and reference image distributions in the Hilbert space $\mathcal{H}$, ensuring distribution alignment.

\textbf{Style Loss.} The style loss is computed using the Gram matrix, which captures the correlations between different feature maps. Let $F^l \in \mathbb{R}^{C_l \times H_l \times W_l}$ represent the feature maps at layer $l$ of the VGG network, where $C_l$, $H_l$, and $W_l$ are the number of channels, height, and width, respectively. The Gram matrix $G^l$ is calculated as:
\begin{equation} 
G^l = F^l (F^l)^T 
\end{equation}
This matrix encodes the style information by capturing the correlations between different channels in the feature maps, effectively preserving the texture and color distribution of the original image. For each layer, we compute the Gram matrix for both the generated image and the reference image. The style loss at each layer is defined as the difference between these Gram matrices, normalized by the size of the layer:
\begin{equation}
\mathcal{L}_{style} = \sum_{l=1}^{L} \frac{1}{4C_l^2 H_l^2 W_l^2} \lVert G^l_{gen} - G^l_{ref} \rVert_F^2
\end{equation}
where $G^l_{gen}$ and $G^l_{ref}$ are the Gram matrices of the generated and reference images at layer $l$, where $l \in \{1, 2, \dots, L\}$ represents the layers of the VGG network, and $\lVert \cdot \rVert_F$ denotes the Frobenius norm.

\textbf{Total Loss.} To balance the content and style losses, we define the total loss function as:
\begin{equation} 
\mathcal{L}_{total} = \mathcal{L}_{recon} + \mathcal{L}_{dis} + \lambda_{style} \mathcal{L}_{style}
\end{equation}
where $\lambda_{style} = 100$ is the weighting factor applied to the style loss. This ensures that the generated image not only maintains content accuracy but also aligns with the reference image's style. By minimizing $\mathcal{L}_{total}$, the VGG-Adapter module effectively reduces color discrepancies and ensures the generated image is visually consistent with the target distribution while maintaining both content and style fidelity.

\begin{figure}[!ht]
    \centering
    \includegraphics[width=0.99\linewidth]{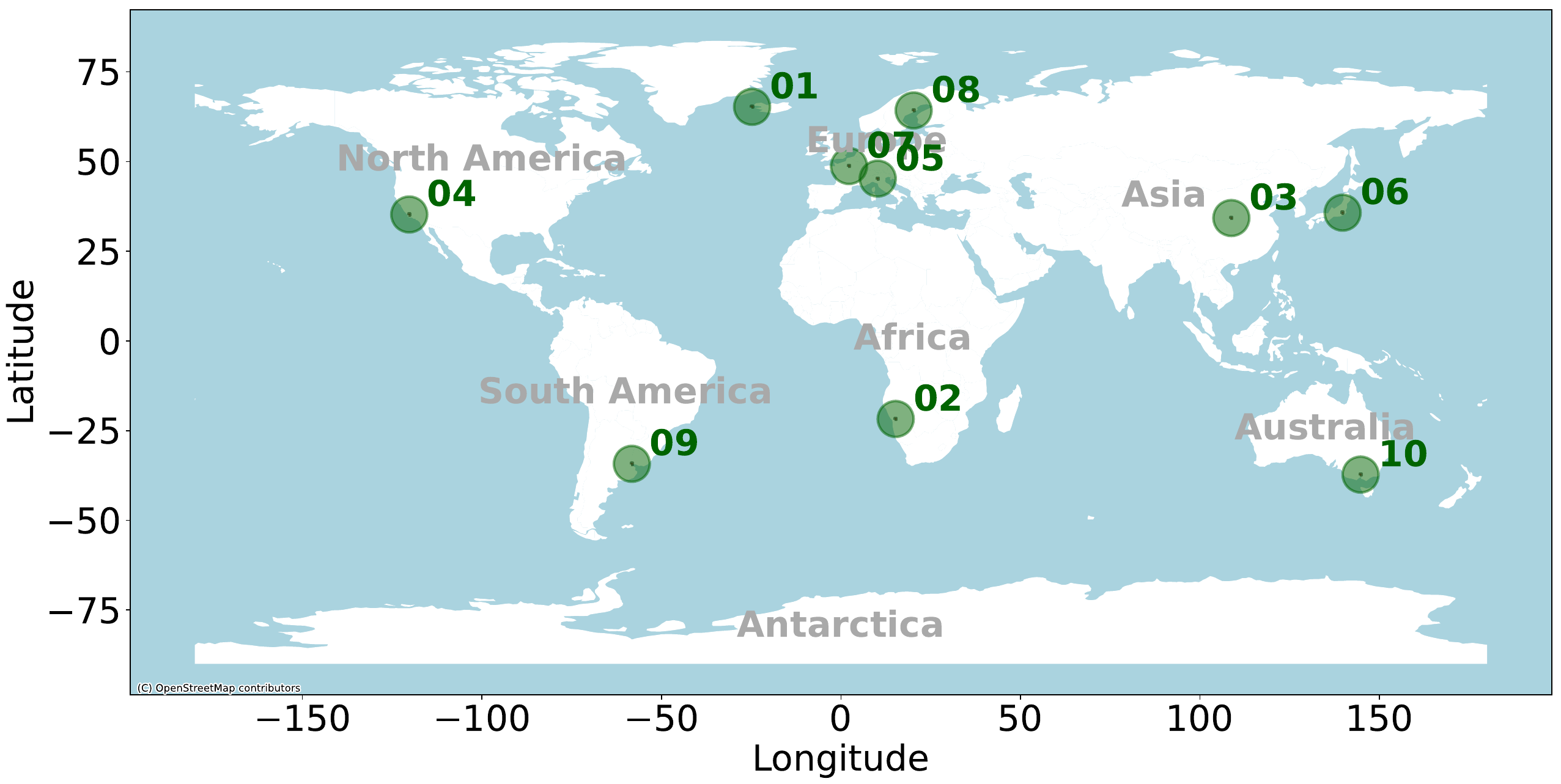}
    \caption{Global distribution of selected regions for Landsat imagery. The imagery was downloaded using Google Earth Engine (GEE) from various countries, ensuring diverse geographical coverage for comprehensive analysis.}
    \label{fig:site_map}
\end{figure}

\begin{table*}
    \centering
    \caption{Data Information. Notes: cloud cover $\leq$ 1\%. The regions were randomly selected, but it was required to ensure that there were available Landsat-8 images within these areas.}
    \resizebox{0.9\textwidth}{!}{
    \begin{tabular}{ccccc}
        \hline
        \textbf{No.} & \textbf{Country / Region} & \textbf{Coordinates (Center Point)} & \textbf{Latitude Range} & \textbf{Longitude Range} \\
        \hline
        01 & New Zealand - Near Christchurch & (-24.75, 65.25) & 65.0 - 65.5 & -25.0 - -24.5 \\
        02 & Namibia - Near Windhoek & (15.25, -21.75) & -22.0 - -21.5 & 15.0 - 15.5 \\
        03 & China - Near Xi'an, Shaanxi Province & (108.75, 34.25) & 34.0 - 34.5 & 108.5 - 109.0 \\
        04 & United States - California & (-120.25, 35.25) & 35.0 - 35.5 & -120.5 - -120.0 \\
        05 & Italy - Northern Region & (10.25, 45.25) & 45.0 - 45.5 & 10.0 - 10.5 \\
        06 & Japan - Near Tokyo & (139.75, 35.75) & 35.5 - 36.0 & 139.5 - 140.0 \\
        07 & France - Near Paris & (2.25, 48.75) & 48.5 - 49.0 & 2.0 - 2.5 \\
        08 & United States - Near Seattle, Washington & (20.25, 64.25) & 64.0 - 64.5 & 20.0 - 20.5 \\
        09 & Argentina - Buenos Aires & (-58.25, -34.25) & -34.5 - -34.0 & -58.5 - -58.0 \\
        10 & Australia - Near Melbourne & (144.75, -37.25) & -37.5 - -37.0 & 144.5 - 145.0 \\
        \hline
    \end{tabular}
    }
    \label{table:data_info}
\end{table*}

\begin{figure*}[!ht]
    \centering
    \begin{minipage}{0.14\textwidth}
        \centering \textbf{DEM}
        \includegraphics[width=\textwidth]{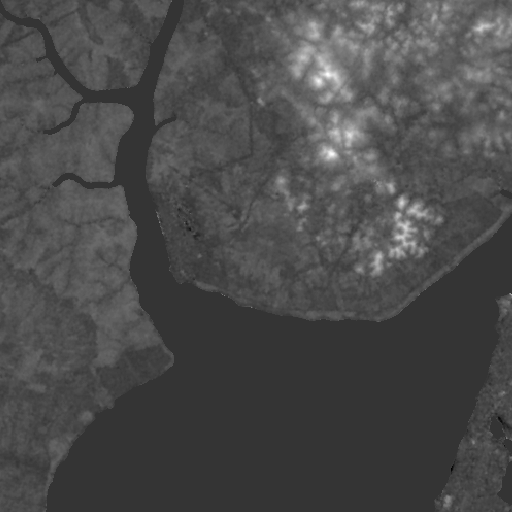}
    \end{minipage}
    \hspace{-3pt}
    \begin{minipage}{0.14\textwidth}
        \centering \textbf{SD}
        \includegraphics[width=\textwidth]{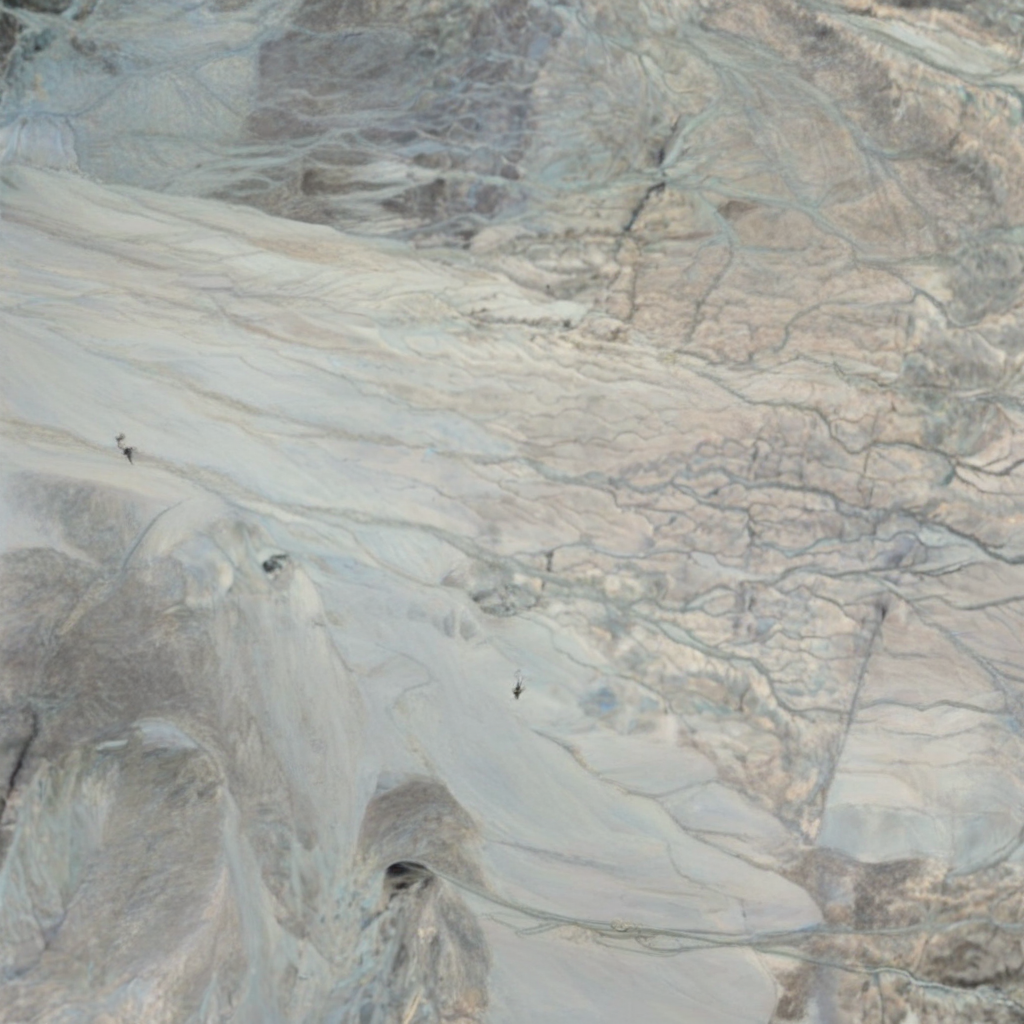}
    \end{minipage}
    \hspace{-3pt}
    \begin{minipage}{0.14\textwidth}
        \centering \textbf{Palette}
        \includegraphics[width=\textwidth]{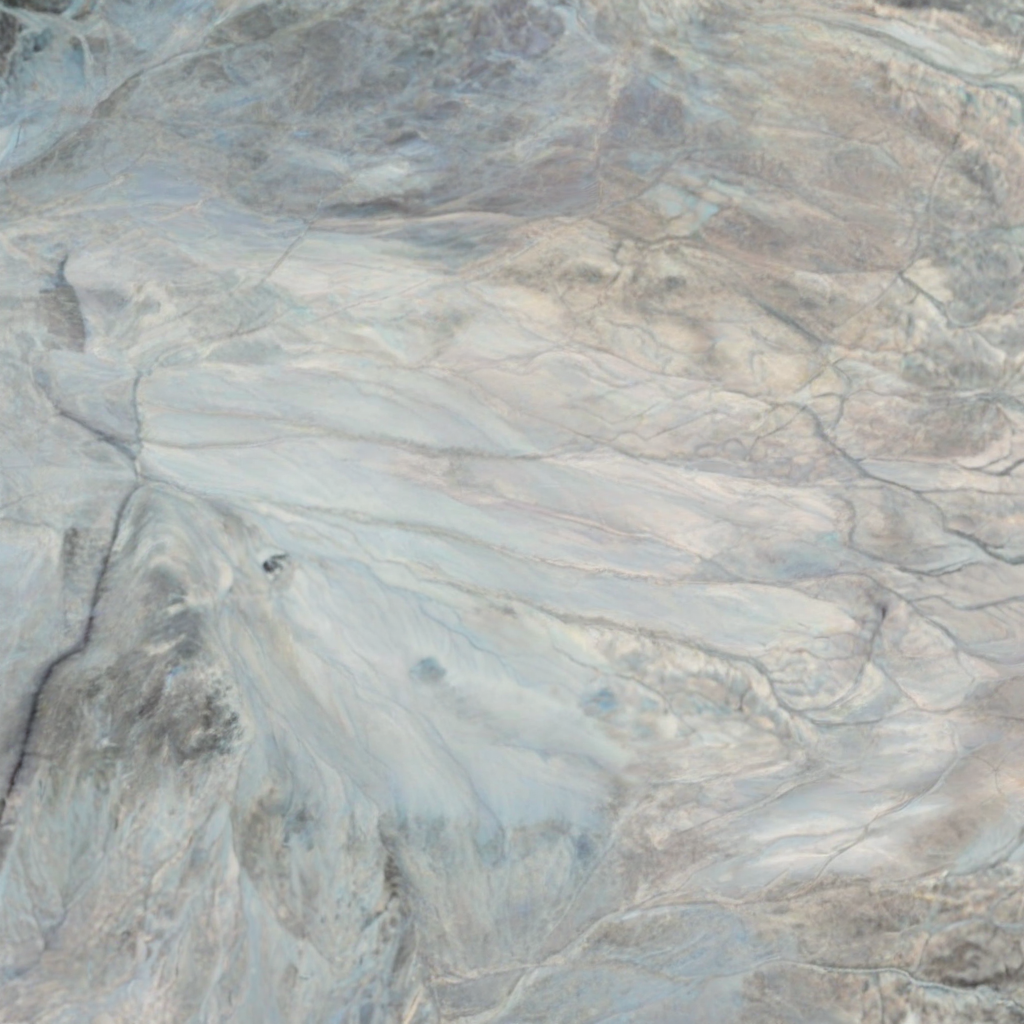}
    \end{minipage}
    \hspace{-3pt}
    \begin{minipage}{0.14\textwidth}
        \centering \textbf{LaMa}
        \includegraphics[width=\textwidth]{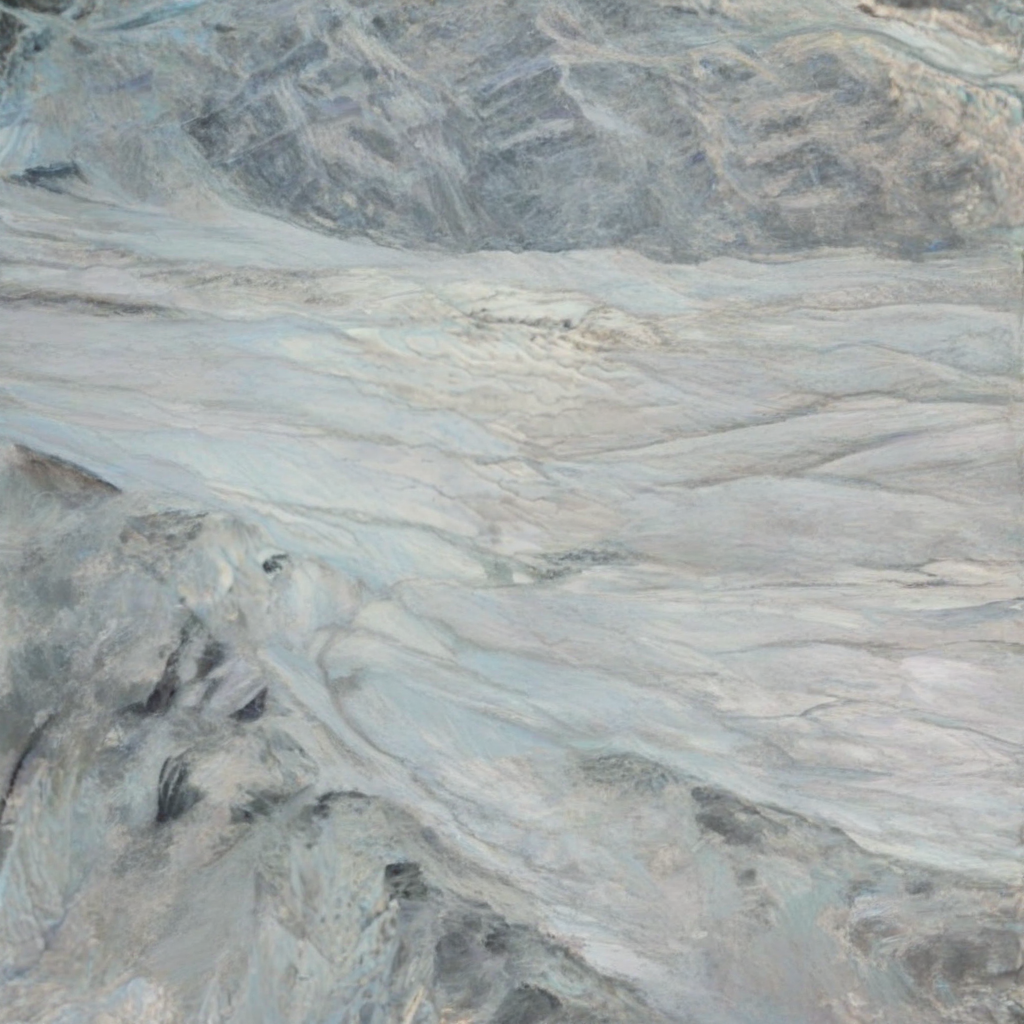}
    \end{minipage}
    \hspace{-3pt}
    \begin{minipage}{0.14\textwidth}
        \centering \textbf{\textit{SatelliteMaker}}
        \includegraphics[width=\textwidth]{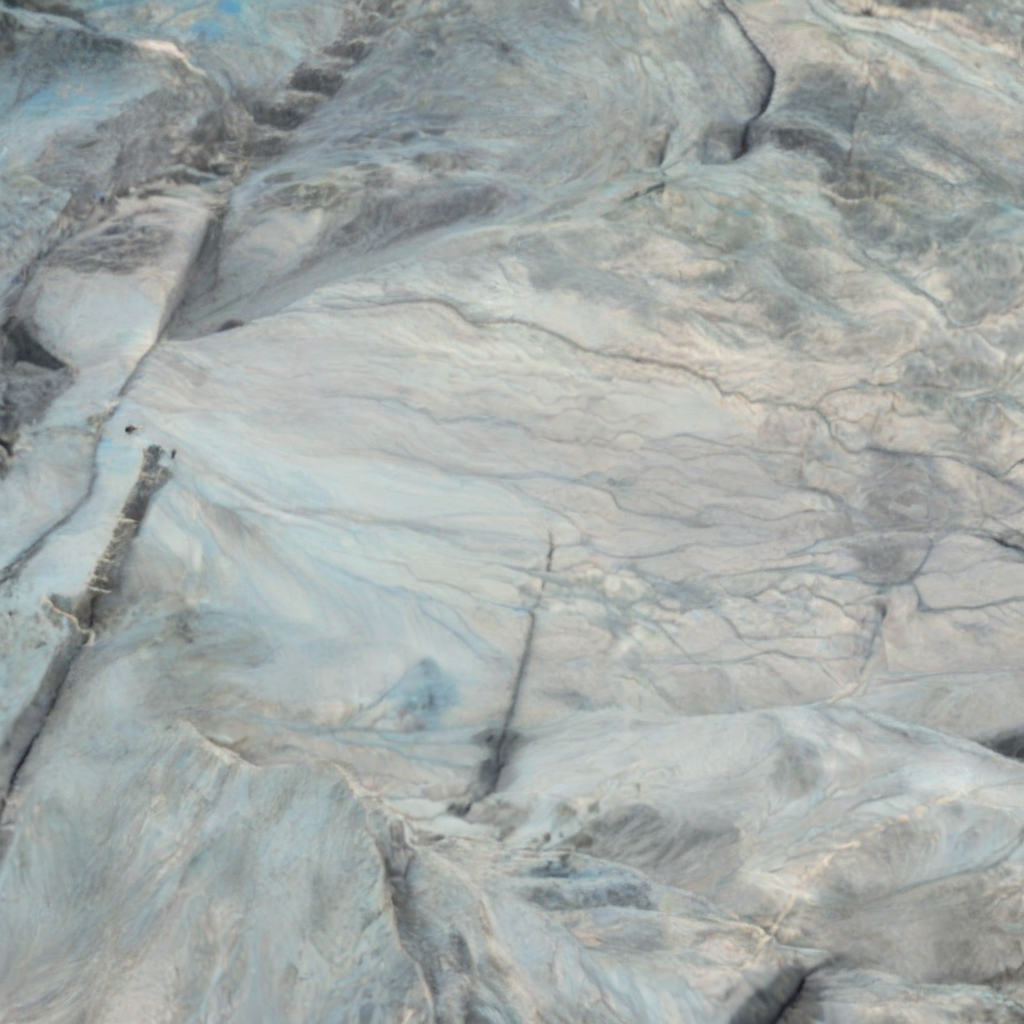}
    \end{minipage}
    \hspace{-3pt}
    \begin{minipage}{0.14\textwidth}
        \centering \textbf{Groundtruth}
        \includegraphics[width=\textwidth]{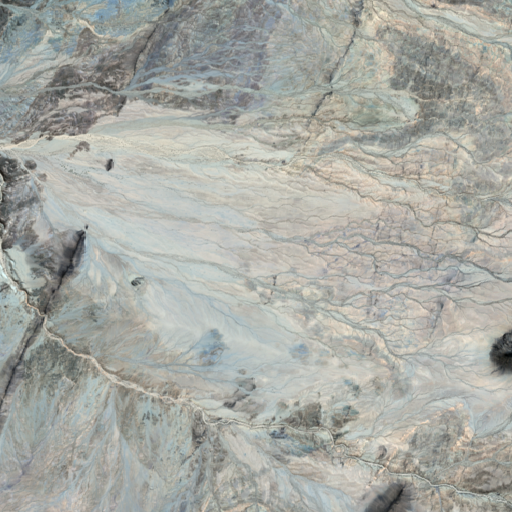}
    \end{minipage}
    \hspace{-3pt}
    \\
    \centering \texttt{Prompt: $<$SatelliteMaker$>$ 02, Namibia, Blue / Green / Red / NIR}
    \\

    \begin{minipage}{0.14\textwidth}
        \includegraphics[width=\textwidth]{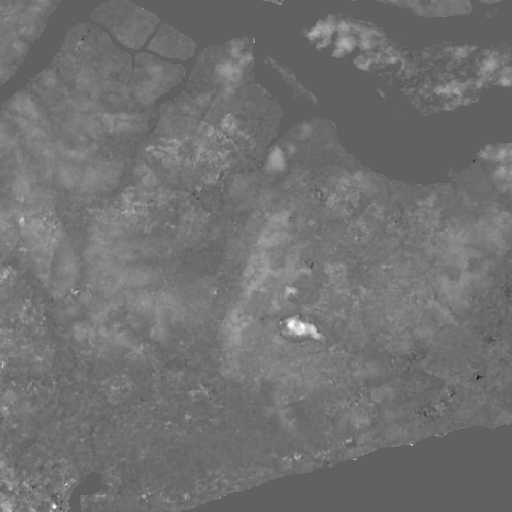}
    \end{minipage}
    \hspace{-3pt}
    \begin{minipage}{0.14\textwidth}
        \includegraphics[width=\textwidth]{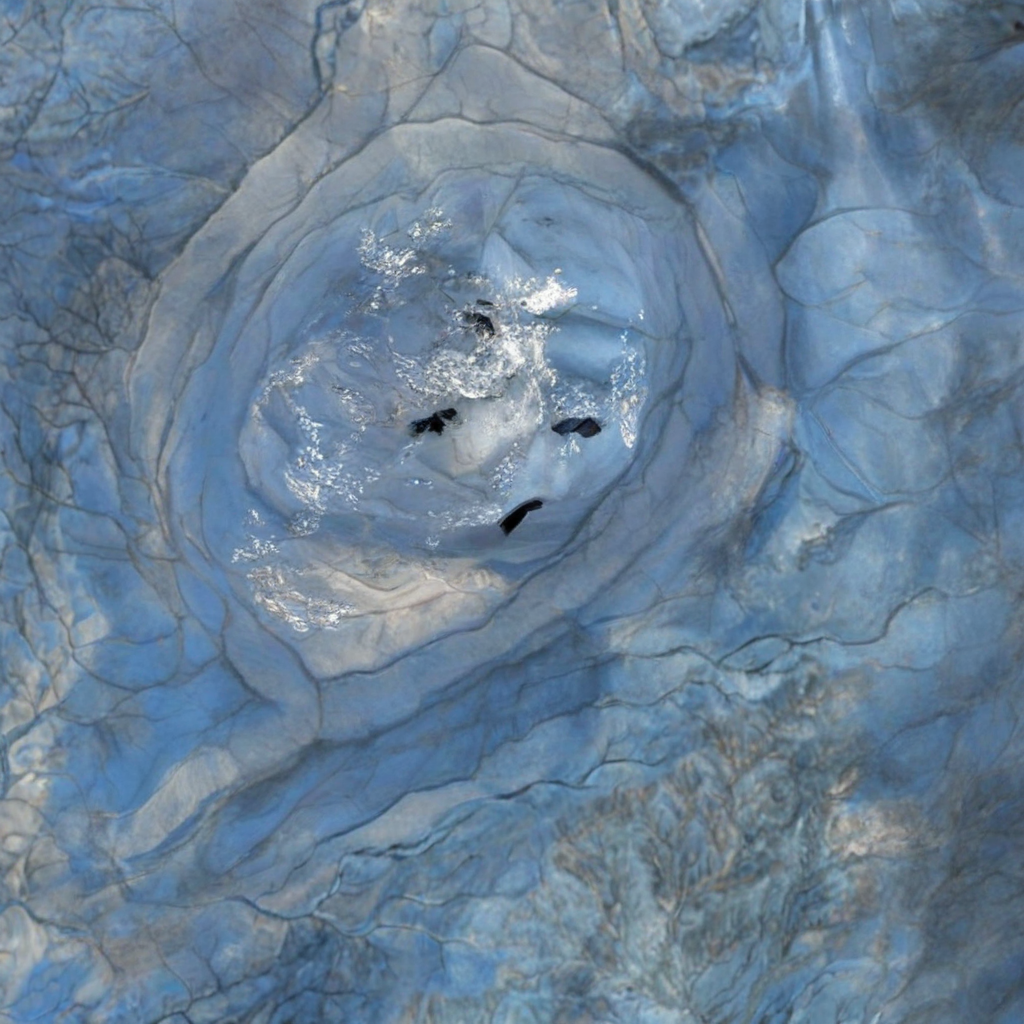}
    \end{minipage}
    \hspace{-3pt}
    \begin{minipage}{0.14\textwidth}
        \includegraphics[width=\textwidth]{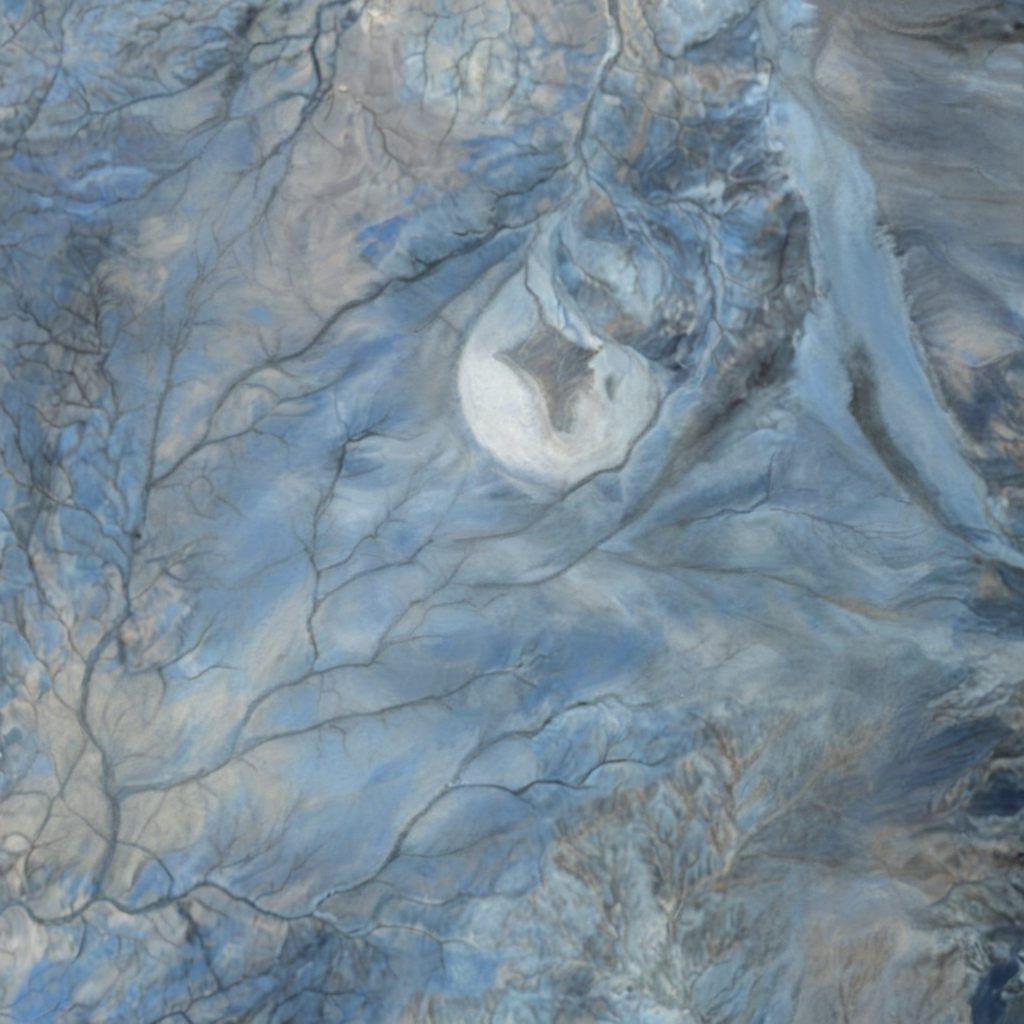}
    \end{minipage}
    \hspace{-3pt}
    \begin{minipage}{0.14\textwidth}
        \includegraphics[width=\textwidth]{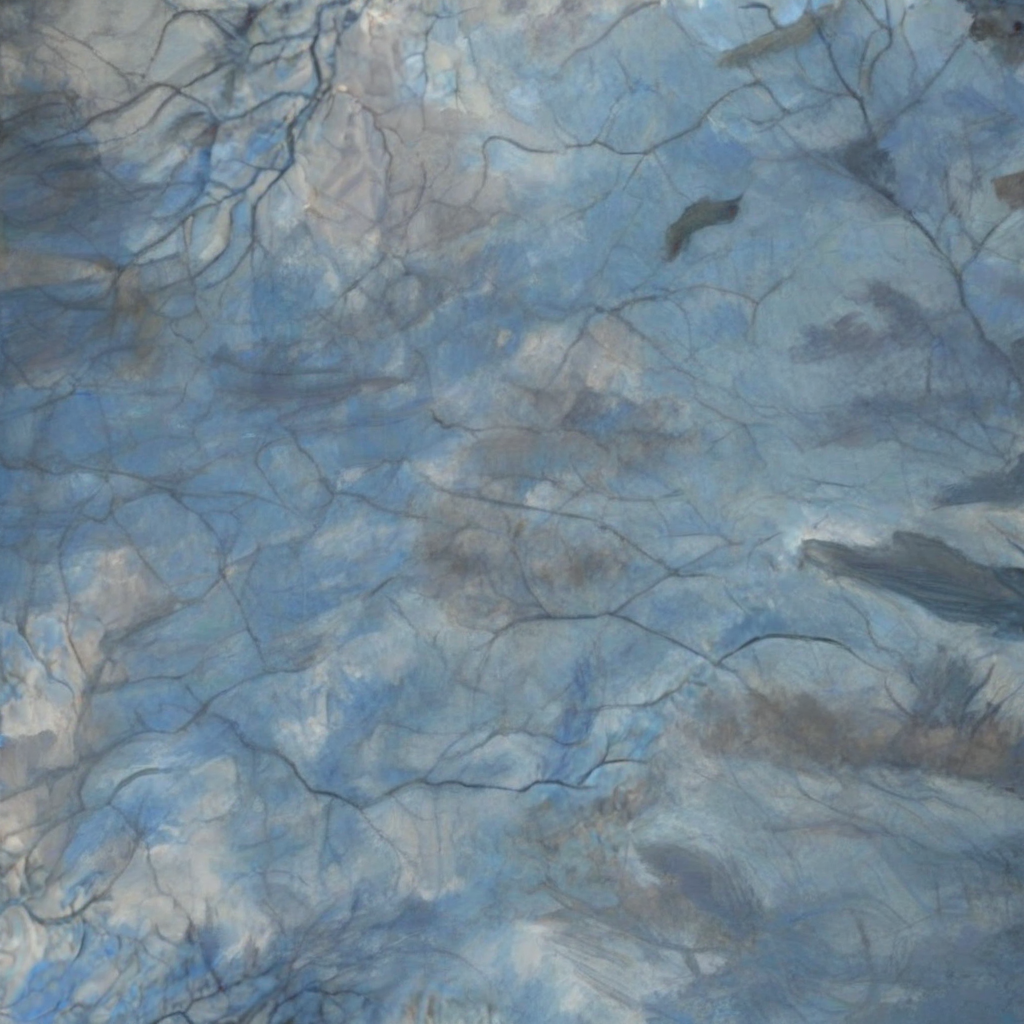}
    \end{minipage}
    \hspace{-3pt}
    \begin{minipage}{0.14\textwidth}
        \includegraphics[width=\textwidth]{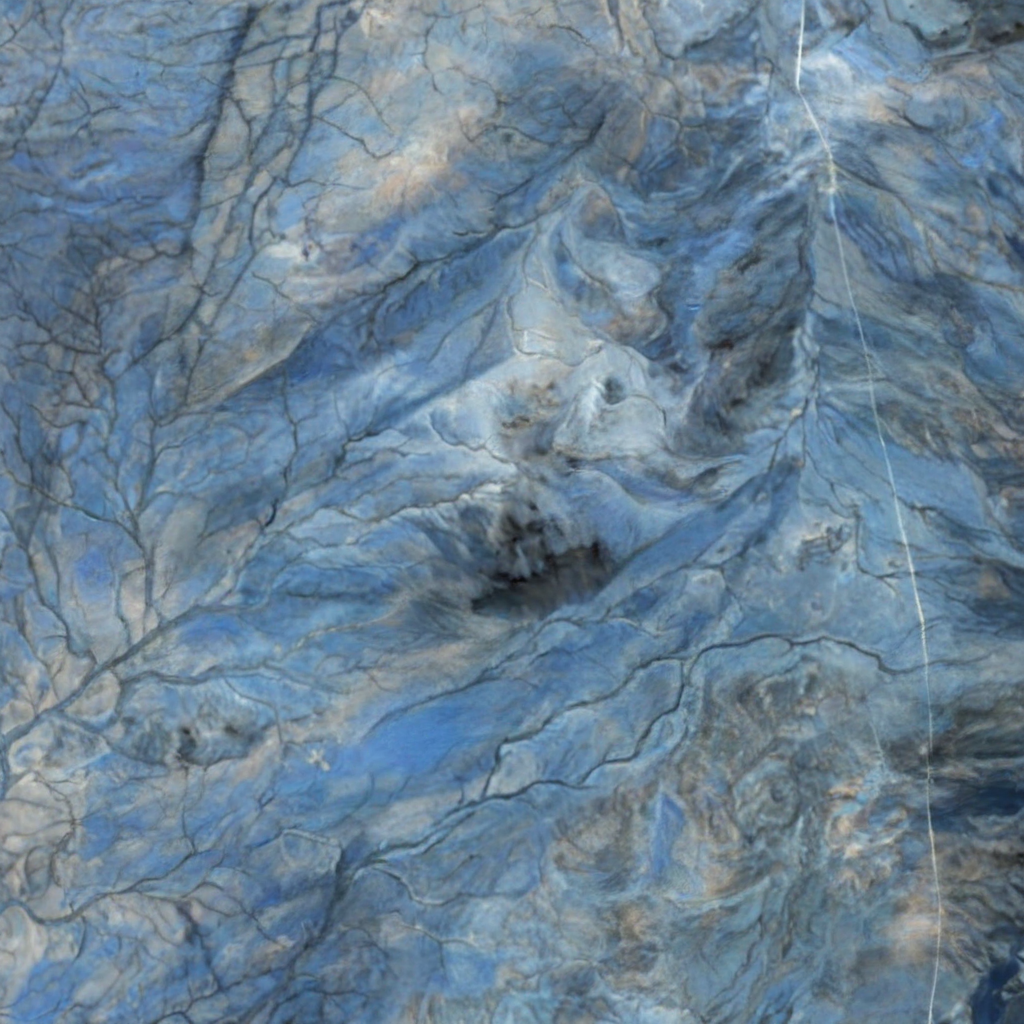}
    \end{minipage}
    \hspace{-3pt}
    \begin{minipage}{0.14\textwidth}
        \includegraphics[width=\textwidth]{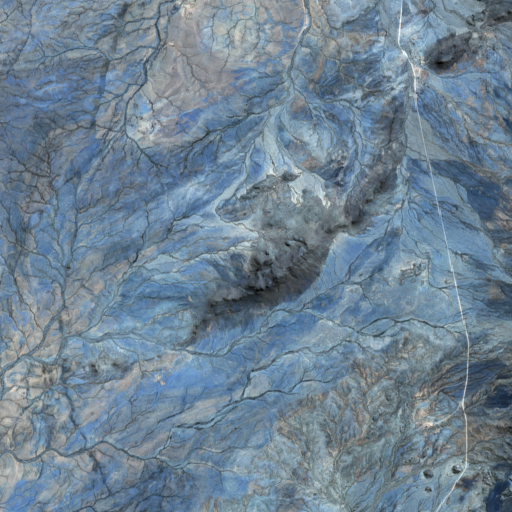}
    \end{minipage}
    \hspace{-3pt}
    \\
    \centering \texttt{Prompt: $<$SatelliteMaker$>$ 02, Namibia, Blue / Green / Red / NIR}
    \\


    \begin{minipage}{0.14\textwidth}
        \includegraphics[width=\textwidth]{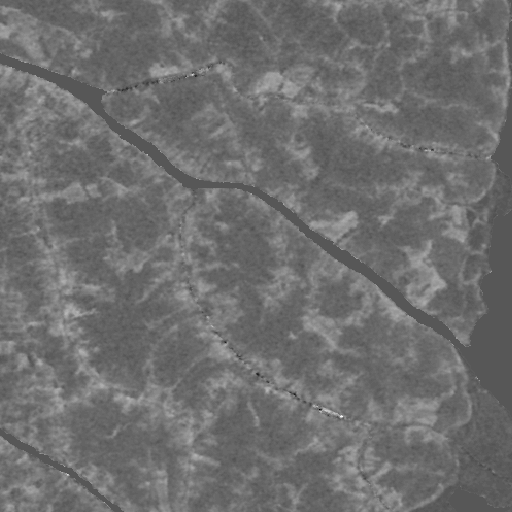}
    \end{minipage}
    \hspace{-3pt}
    \begin{minipage}{0.14\textwidth}
        \includegraphics[width=\textwidth]{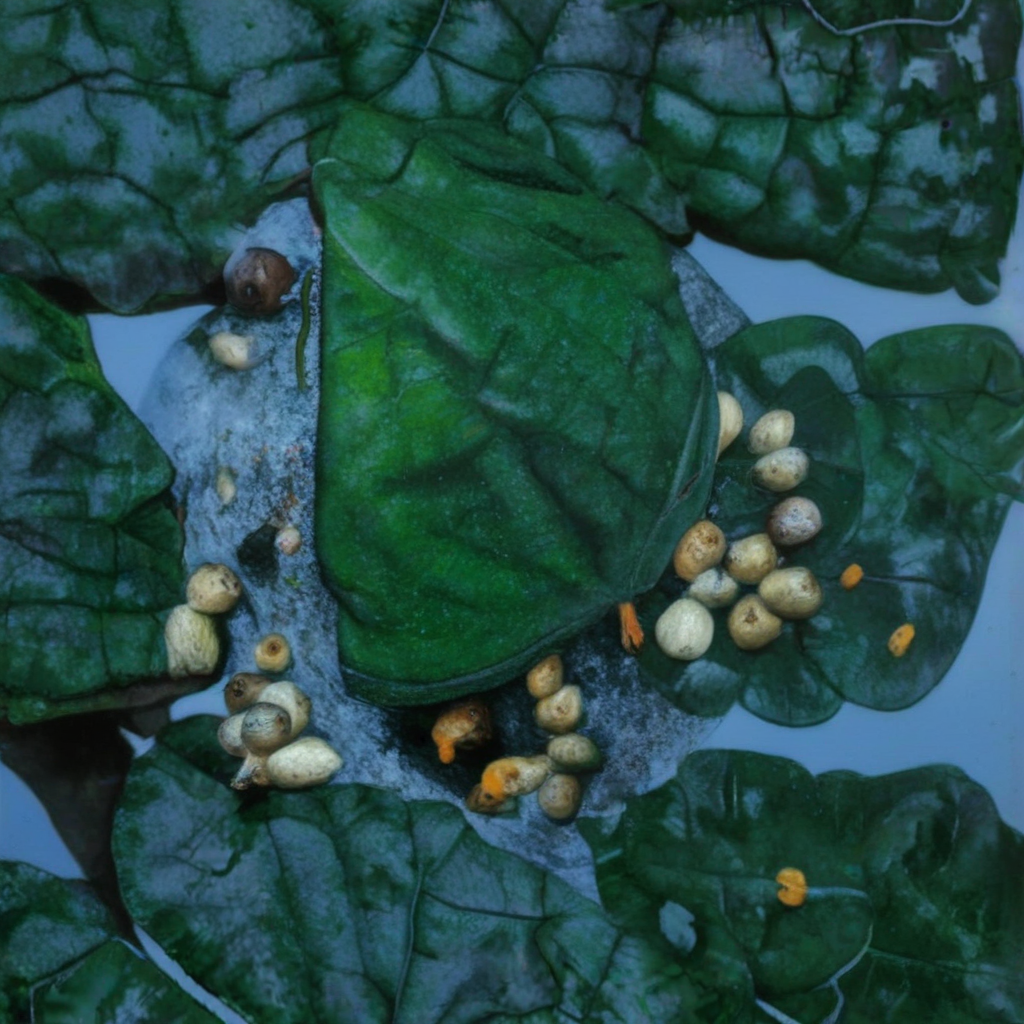}
    \end{minipage}
    \hspace{-3pt}
    \begin{minipage}{0.14\textwidth}
        \includegraphics[width=\textwidth]{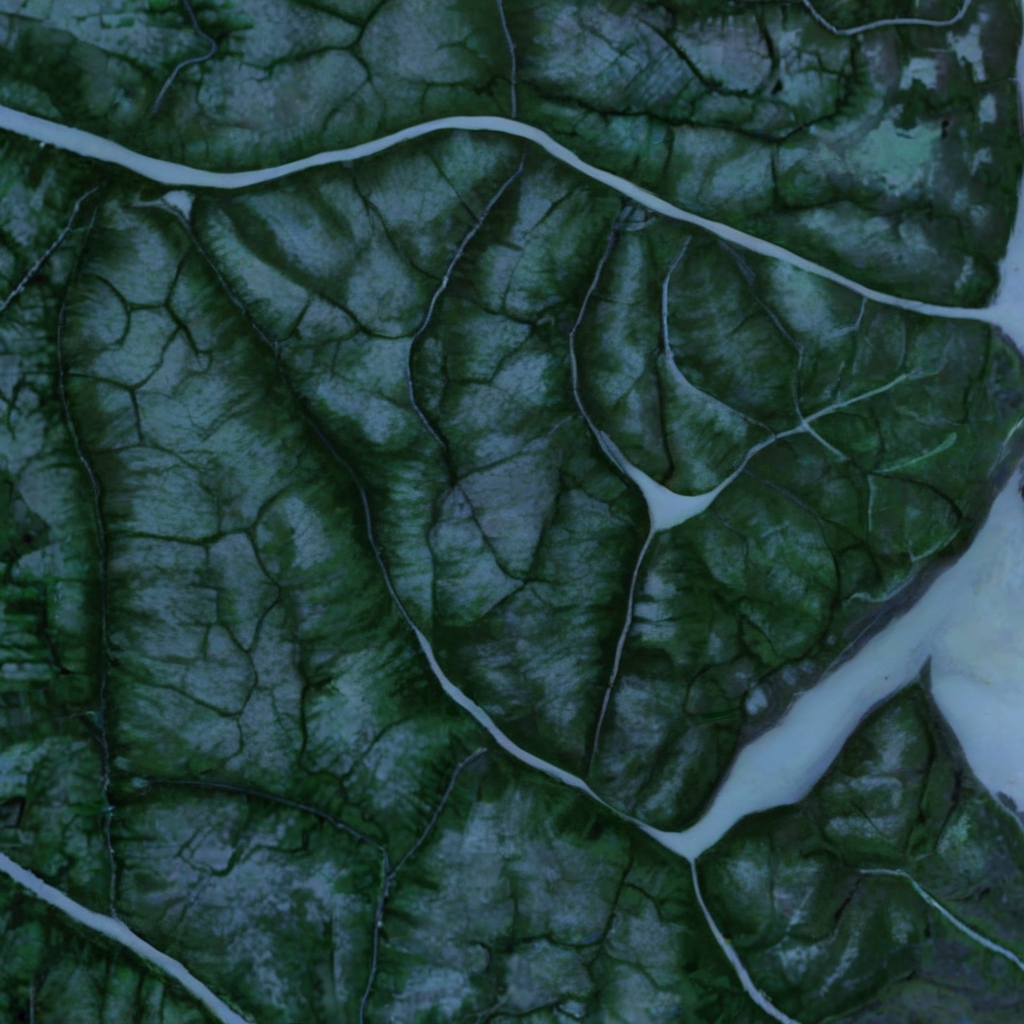}
    \end{minipage}
    \hspace{-3pt}
    \begin{minipage}{0.14\textwidth}
        \includegraphics[width=\textwidth]{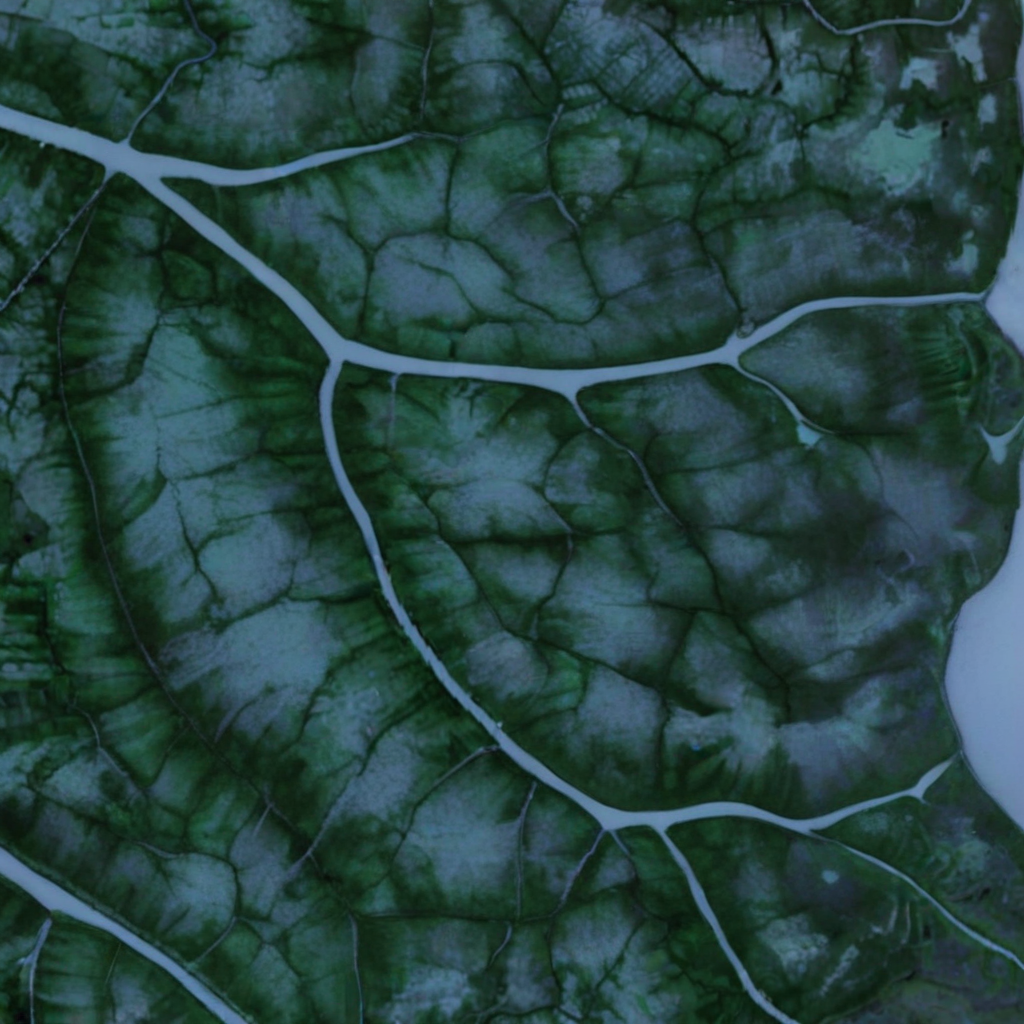}
    \end{minipage}
    \hspace{-3pt}
    \begin{minipage}{0.14\textwidth}
        \includegraphics[width=\textwidth]{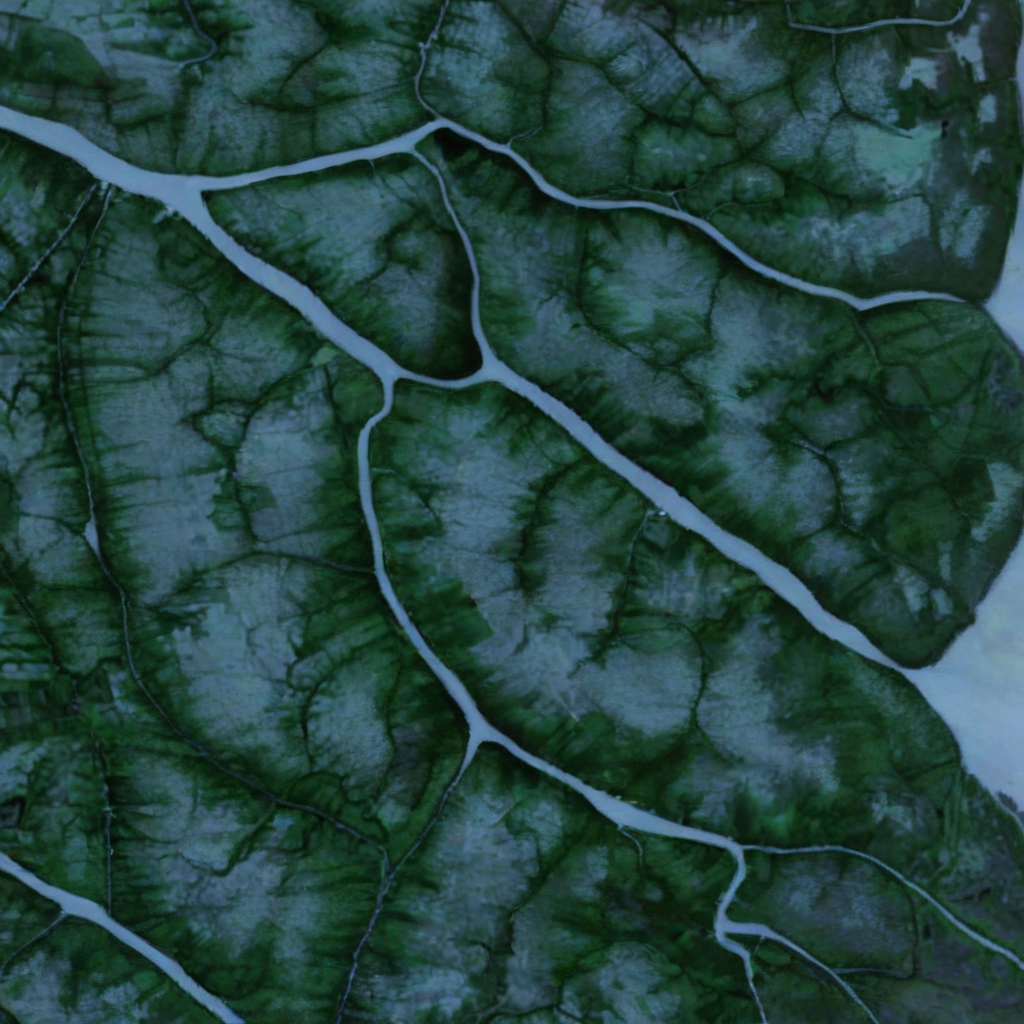}
    \end{minipage}
    \hspace{-3pt}
    \begin{minipage}{0.14\textwidth}
        \includegraphics[width=\textwidth]{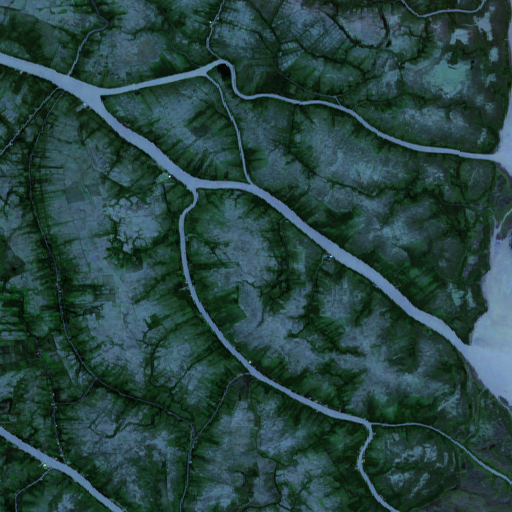}
    \end{minipage}
    \hspace{-3pt}
    \\
    \centering \texttt{Prompt: $<$SatelliteMaker$>$ 09, Argentina, Blue / Green / Red / NIR}
    \\

    \begin{minipage}{0.14\textwidth}
        \includegraphics[width=\textwidth]{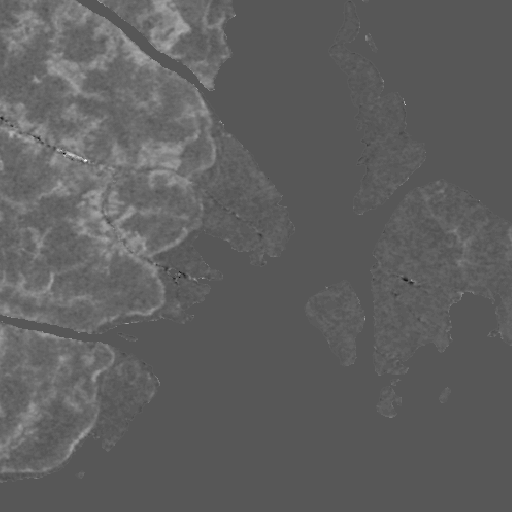}
    \end{minipage}
    \hspace{-3pt}
    \begin{minipage}{0.14\textwidth}
        \includegraphics[width=\textwidth]{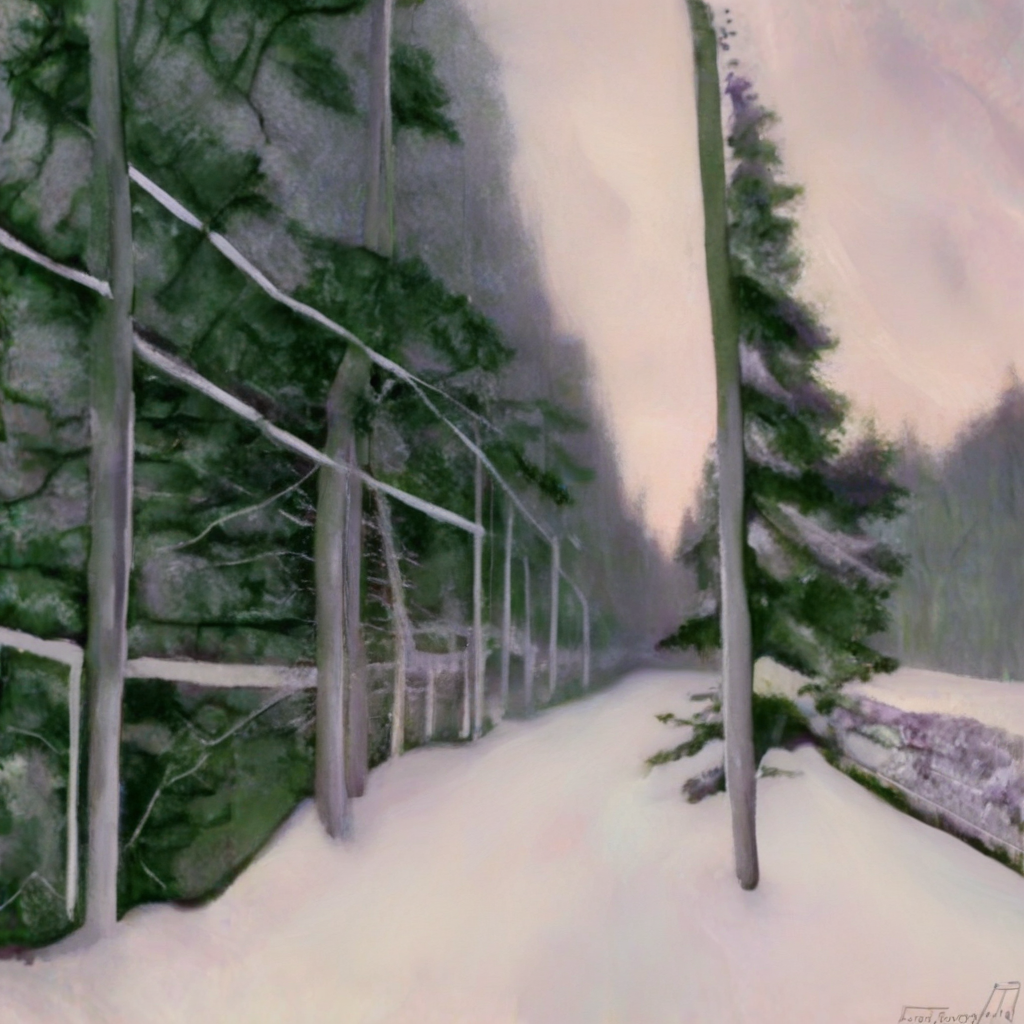}
    \end{minipage}
    \hspace{-3pt}
    \begin{minipage}{0.14\textwidth}
        \includegraphics[width=\textwidth]{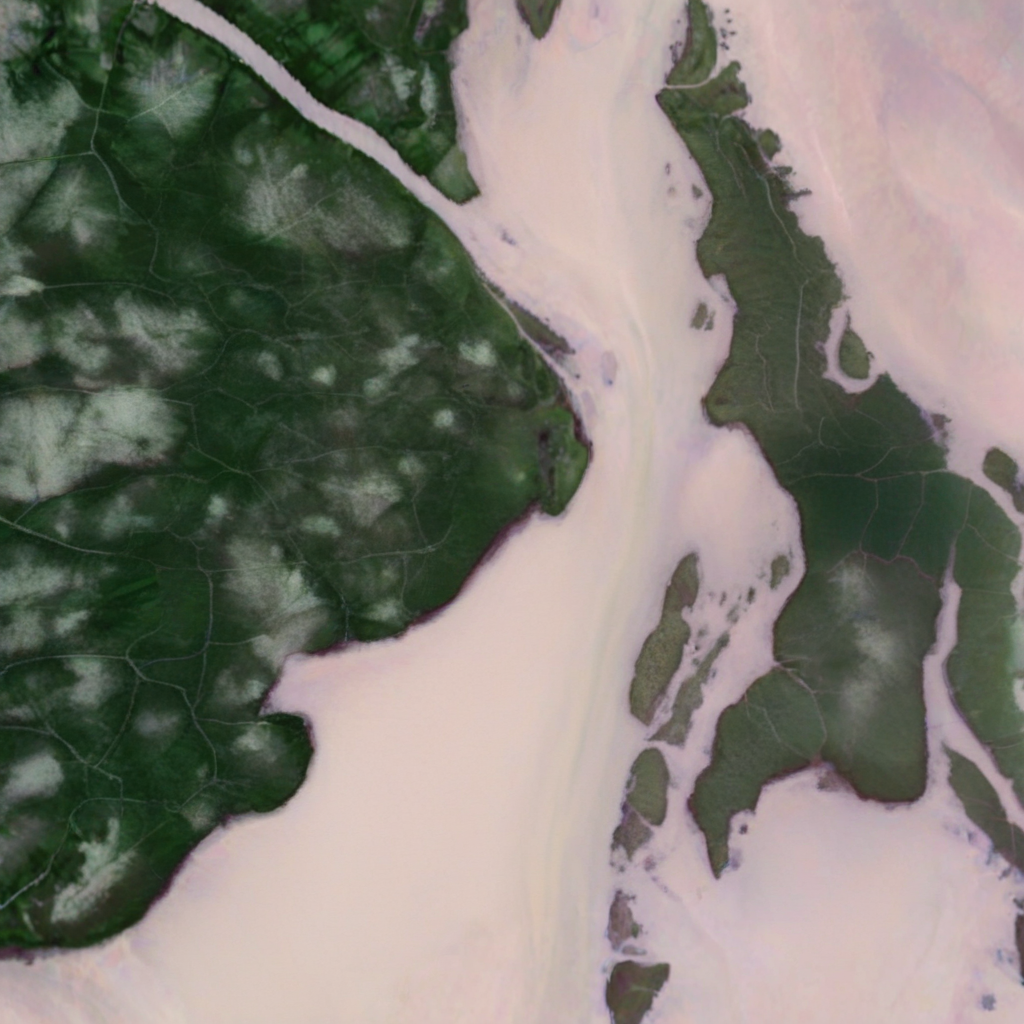}
    \end{minipage}
    \hspace{-3pt}
    \begin{minipage}{0.14\textwidth}
        \includegraphics[width=\textwidth]{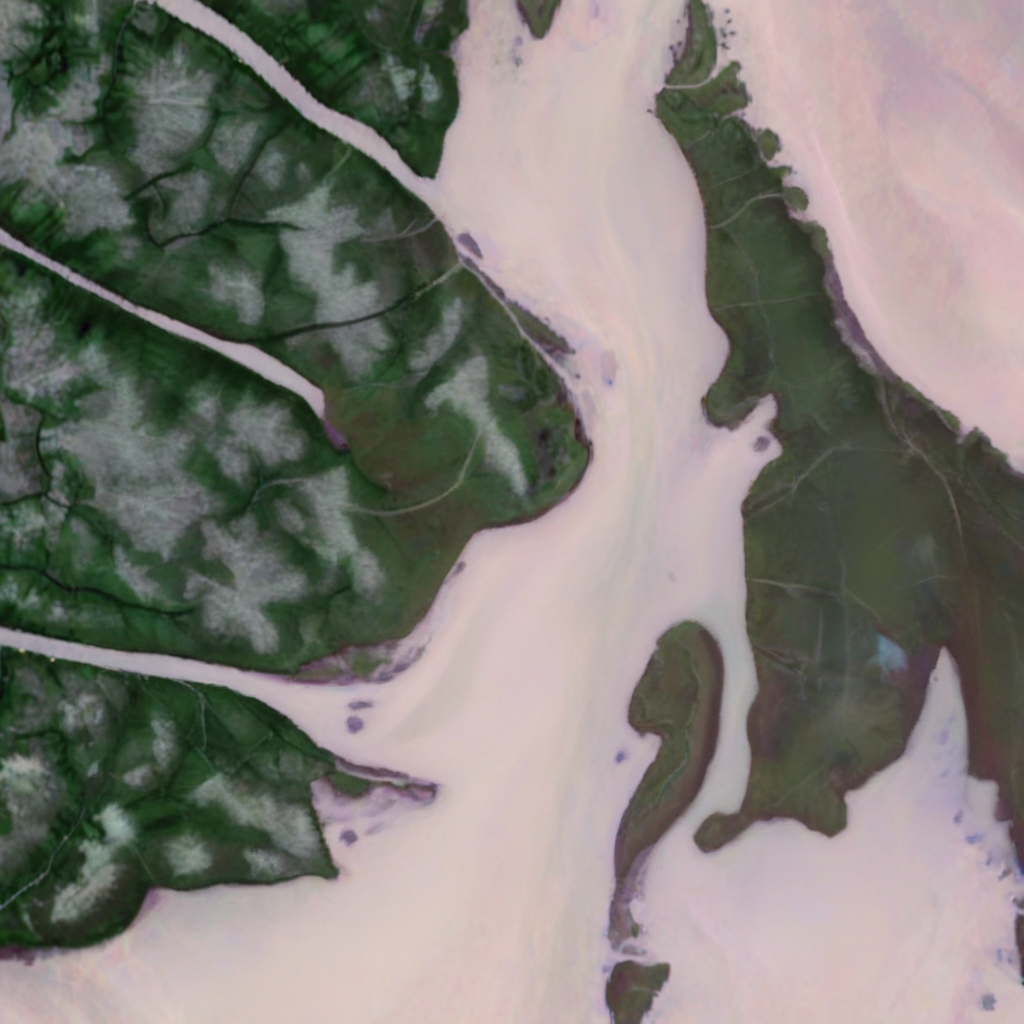}
    \end{minipage}
    \hspace{-3pt}
    \begin{minipage}{0.14\textwidth}
        \includegraphics[width=\textwidth]{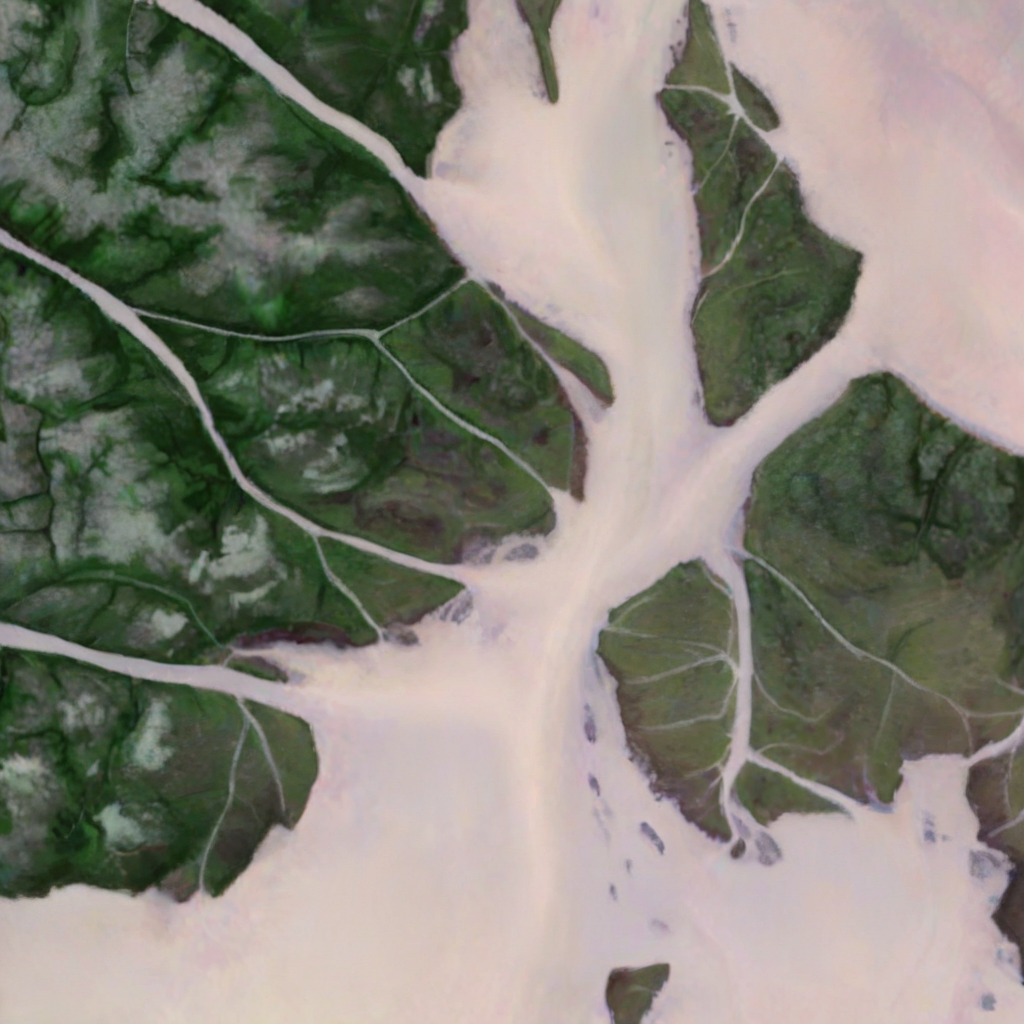}
    \end{minipage}
    \hspace{-3pt}
    \begin{minipage}{0.14\textwidth}
        \includegraphics[width=\textwidth]{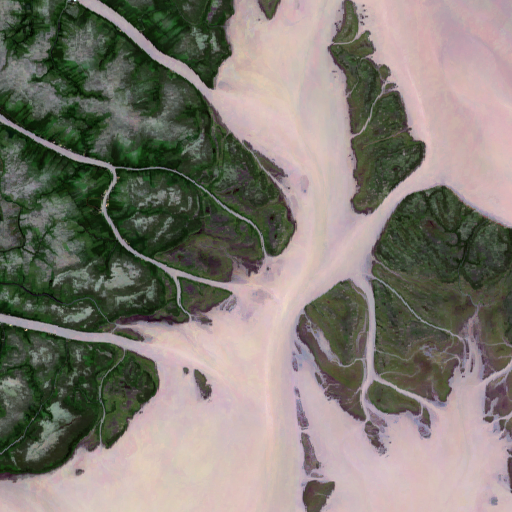}
    \end{minipage}
    \hspace{-3pt}
    \\
    \centering \texttt{Prompt: $<$SatelliteMaker$>$ 09, Argentina, Blue / Green / Red / NIR}
    \\
    \caption{Comparison for Task-1, addressing missing data in specific regions over a fixed time period. }
    \label{fig_compare1}
\end{figure*}

\begin{figure*}[!ht]
    \centering
    \begin{minipage}{0.135\textwidth}
        \centering \textbf{DEM}
        \includegraphics[width=\textwidth]{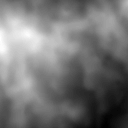}
    \end{minipage}
    \hspace{-3pt}
    \begin{minipage}{0.135\textwidth}
        \centering \textbf{Interpolation}
        \includegraphics[width=\textwidth]{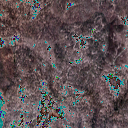}
    \end{minipage}
    \hspace{-3pt}
    \begin{minipage}{0.135\textwidth}
        \centering \textbf{STCNN}
        \includegraphics[width=\textwidth]{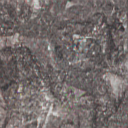}
    \end{minipage}
    \hspace{-3pt}
    \begin{minipage}{0.135\textwidth}
        \centering \textbf{AutoEncoder}
        \includegraphics[width=\textwidth]{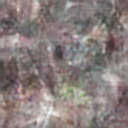}
    \end{minipage}
    \hspace{-3pt}
    \begin{minipage}{0.135\textwidth}
        \centering \textbf{ControlNet}
        \includegraphics[width=\textwidth]{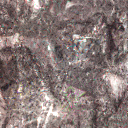}
    \end{minipage}
    \hspace{-3pt}
    \begin{minipage}{0.135\textwidth}
        \centering \textbf{\textit{SatelliteMaker}}
        \includegraphics[width=\textwidth]{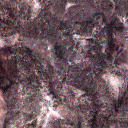}
    \end{minipage}
    \hspace{-3pt}
    \begin{minipage}{0.135\textwidth}
        \centering \textbf{Groundtruth}
        \includegraphics[width=\textwidth]{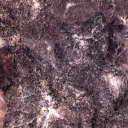}
    \end{minipage}
    \hspace{-3pt}
    \\
    \centering \texttt{Prompt: $<$SatelliteMaker$>$ SQC, 2017-07-02, Day:6, Blue / Green / Red / NIR}
    \\
    \begin{minipage}{0.135\textwidth}
        \includegraphics[width=\textwidth]{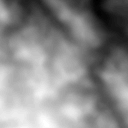}
    \end{minipage}
    \hspace{-3pt}
    \begin{minipage}{0.135\textwidth}
        \includegraphics[width=\textwidth]{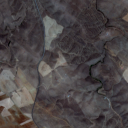}
    \end{minipage}
    \hspace{-3pt}
    \begin{minipage}{0.135\textwidth}
        \includegraphics[width=\textwidth]{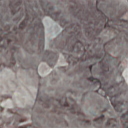}
    \end{minipage}
    \hspace{-3pt}
    \begin{minipage}{0.135\textwidth}
        \includegraphics[width=\textwidth]{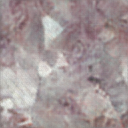}
    \end{minipage}
    \hspace{-3pt}
    \begin{minipage}{0.135\textwidth}
        \includegraphics[width=\textwidth]{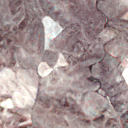}
    \end{minipage}
    \hspace{-3pt}
    \begin{minipage}{0.135\textwidth}
        \includegraphics[width=\textwidth]{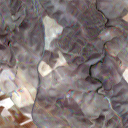}
    \end{minipage}
    \hspace{-3pt}
    \begin{minipage}{0.135\textwidth}
        \includegraphics[width=\textwidth]{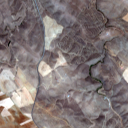}
    \end{minipage}
    \hspace{-3pt}
    \\
    \centering \texttt{Prompt: $<$SatelliteMaker$>$ SQC, 2017-07-07, Day:6, Blue / Green / Red / NIR}
    \\
    \begin{minipage}{0.135\textwidth}
        \includegraphics[width=\textwidth]{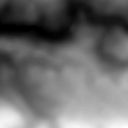}
    \end{minipage}
    \hspace{-3pt}
    \begin{minipage}{0.135\textwidth}
        \includegraphics[width=\textwidth]{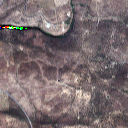}
    \end{minipage}
    \hspace{-3pt}
    \begin{minipage}{0.135\textwidth}
        \includegraphics[width=\textwidth]{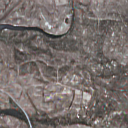}
    \end{minipage}
    \hspace{-3pt}
    \begin{minipage}{0.135\textwidth}
        \includegraphics[width=\textwidth]{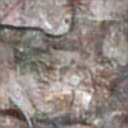}
    \end{minipage}
    \hspace{-3pt}
    \begin{minipage}{0.135\textwidth}
        \includegraphics[width=\textwidth]{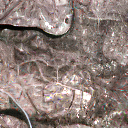}
    \end{minipage}
    \hspace{-3pt}
    \begin{minipage}{0.135\textwidth}
        \includegraphics[width=\textwidth]{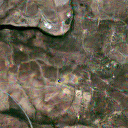}
    \end{minipage}
    \hspace{-3pt}
    \begin{minipage}{0.135\textwidth}
        \includegraphics[width=\textwidth]{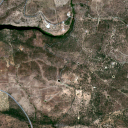}
    \end{minipage}
    \hspace{-3pt}
    \\
    \centering \texttt{Prompt: $<$SatelliteMaker$>$ SQC, 2017-07-07, Day:1, Blue / Green / Red / NIR}
    \\
    \begin{minipage}{0.135\textwidth}
        \includegraphics[width=\textwidth]{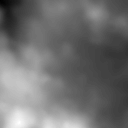}
    \end{minipage}
    \hspace{-3pt}
    \begin{minipage}{0.135\textwidth}
        \includegraphics[width=\textwidth]{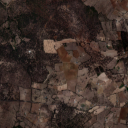}
    \end{minipage}
    \hspace{-3pt}
    \begin{minipage}{0.135\textwidth}
        \includegraphics[width=\textwidth]{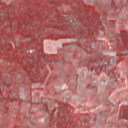}
    \end{minipage}
    \hspace{-3pt}
    \begin{minipage}{0.135\textwidth}
        \includegraphics[width=\textwidth]{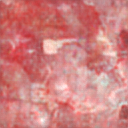}
    \end{minipage}
    \hspace{-3pt}
    \begin{minipage}{0.135\textwidth}
        \includegraphics[width=\textwidth]{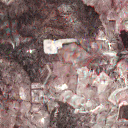}
    \end{minipage}
    \hspace{-3pt}
    \begin{minipage}{0.135\textwidth}
        \includegraphics[width=\textwidth]{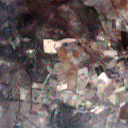}
    \end{minipage}
    \hspace{-3pt}
    \begin{minipage}{0.135\textwidth}
        \includegraphics[width=\textwidth]{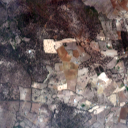}
    \end{minipage}
    \hspace{-3pt}
    \\
    \centering \texttt{Prompt: $<$SatelliteMaker$>$ SQC, 2017-07-17, Day:6, Blue / Green / Red / NIR}
    \\
    \caption{Comparison for Task-2 using the EarthNet2021 dataset with selected missing data days. The above images are the results after brightness adjustment and gamma correction, with a coefficient of 1.2. The original image was used for calculating the  evaluation metrics.}
    \label{fig_compare2}
\end{figure*}

\section{Experiment}
\subsection{Dataset Description}

\textbf{Task-1: Landsat-8 Multi-Region Dataset}. 
We utilize Landsat-8 imagery and DEM data from 10 globally distributed regions, spanning diverse climates and landscapes, including urban and natural environments. Table \ref{table:data_info} lists the regions, and Figure \ref{fig:site_map} illustrates their spatial distribution.
All images have cloud cover below 1\%, filtered using Google Earth Engine (GEE). The 30m resolution images are divided into 512×512 pixel tiles with 50\% overlap to enhance data diversity and model continuity. The complete data processing workflow, including GEE scripts, is provided in Appendix \ref{appendix_geecode}.

\textbf{Task-2: EarthNet2021 Spatiotemporal Dataset}.  
The EarthNet2021 dataset \cite{requena2021earthnet2021} consists of over 200,000 Sentinel-2 optical and multispectral images, supporting earth surface forecasting, environmental monitoring, and image reconstruction. 
We filter out images where low-quality pixels (e.g., clouds, shadows) exceed 5\%. The dataset is divided into IID-test and OOD-test sets, following official benchmarks for evaluating model performance under in-distribution and out-of-distribution scenarios.

\textbf{Masking Strategy}.  
To assess the model's reconstruction capability, we apply the following masking strategy:  
(1) {Cloud Removal}: All cloud-contaminated regions are excluded to ensure data quality.  
(2) {Random Masking}: 10\%-50\% of the remaining high-quality pixels are randomly masked to simulate data loss.  
(3) {Reproducibility}: The masked regions are publicly annotated for standardized evaluation.  
This strategy is consistently applied to both Task-1 and Task-2, ensuring comparability across datasets.

\begin{figure*}[ht]
    \centering
    \begin{minipage}[b]{0.33\textwidth}
        \centering
        \includegraphics[width=\linewidth]{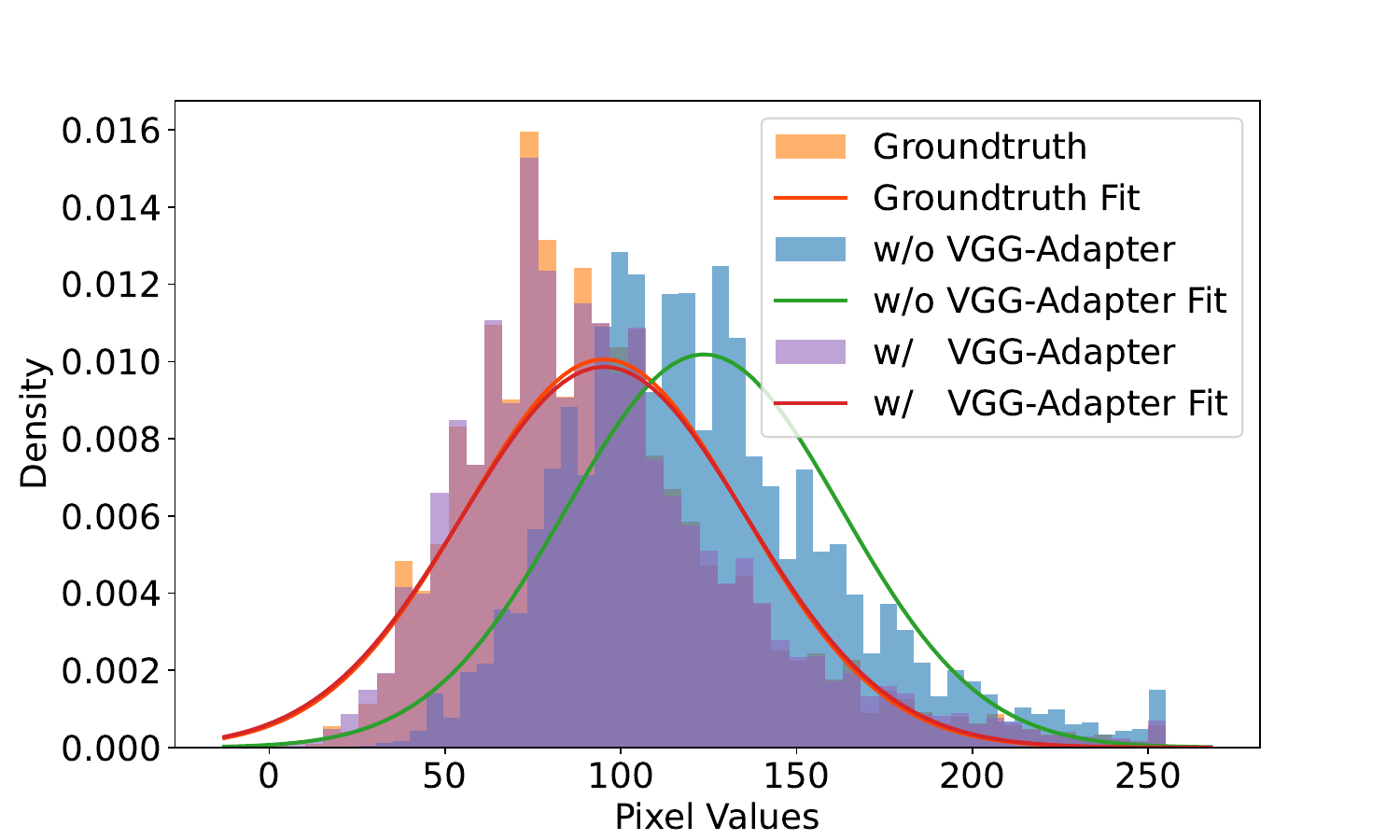}
        \caption{Ablation study on the impact of adding the VGG-Adapter.}
        \label{fig_ablation_density}
    \end{minipage}
    \hfill
    \begin{minipage}[b]{0.64\textwidth}
        \centering
        \begin{subfigure}[b]{0.49\linewidth}
            \centering
            \includegraphics[width=\linewidth]{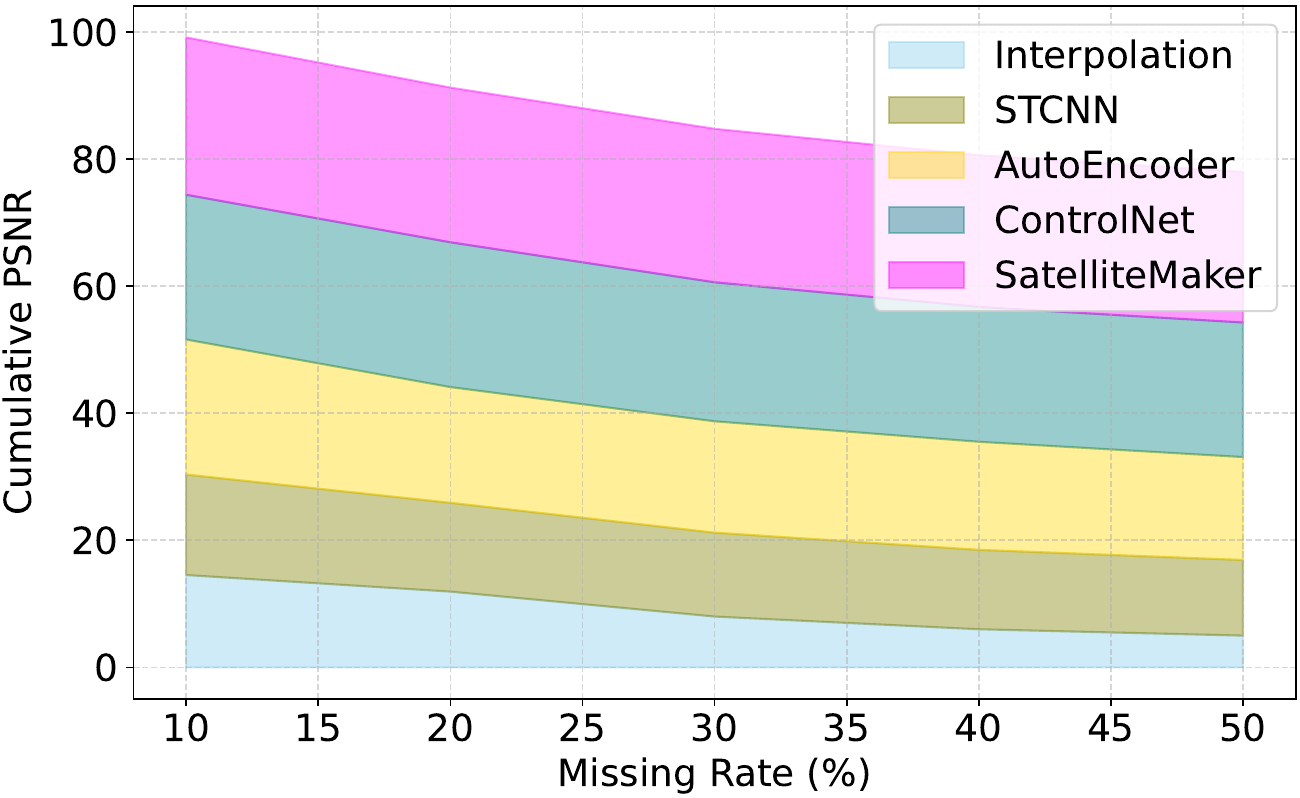}
            \caption{PSNR}
            \label{fig_ablation_psnr}
        \end{subfigure}
        \hfill
        \begin{subfigure}[b]{0.49\linewidth}
            \centering
            \includegraphics[width=\linewidth]{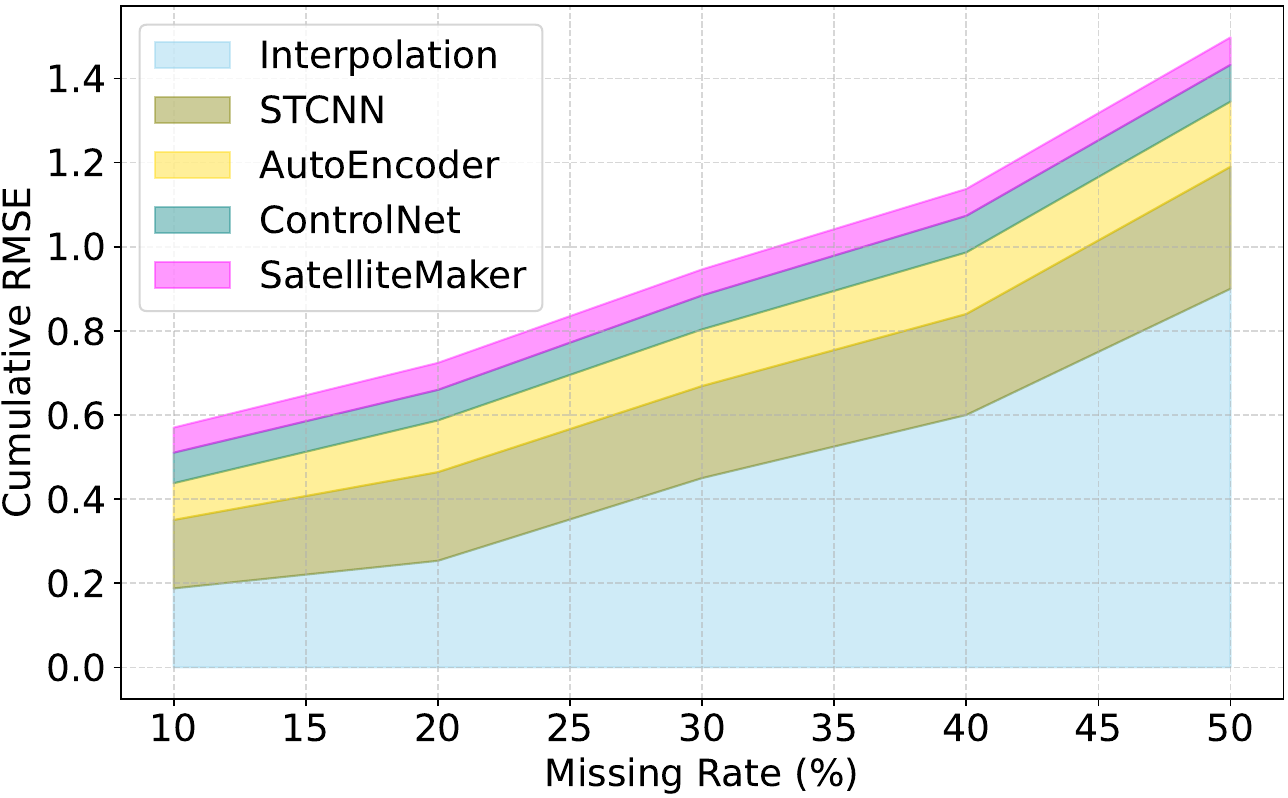}
            \caption{RMSE}
            \label{fig_ablation_rmse}
        \end{subfigure}
        \caption{Comparison of different methods using PSNR and RMSE.}
        \label{fig_ablation_combined}
    \end{minipage}
\end{figure*}

\subsection{Experimental Settings}





The experiments were conducted on a single NVIDIA A100 GPU with 80 GB of memory. The parameter settings are as follows: the initial learning rate was set to $5 \times 10^{-5}$. The image resolution was 512, the detection resolution was 384, and DDIM steps were 50. The $strength$ was 0.9, the $scale$ was 1.0, and $eta$ was 1.0. In the comparison experiments, the STCNN training settings included a batch size of 16, a learning rate of $10^{-4}$, and training for 100 epochs. Regarding data selection, we split the data into 80\% for training and 20\% for testing.


\subsection{Evaluation Metrics}


To evaluate the performance of the proposed \textit{SatelliteMaker} model, we utilize five commonly used metrics: Root Mean Square Error (RMSE) \cite{willmott2005advantages}, Mean Absolute Error (MAE) \cite{chai2014root}, Peak Signal-to-Noise Ratio (PSNR) \cite{hore2010image}, Structural Similarity Index Measure (SSIM) \cite{wang2004image}, and Learned Perceptual Image Patch Similarity (LPIPS) \cite{zhang2018unreasonable}. RMSE measures the pixel-wise differences between the reconstructed and true images, with lower values indicating higher accuracy. MAE calculates the average absolute differences, with smaller values representing better reconstruction accuracy. PSNR assesses the overall quality of image reconstruction, where higher values denote better fidelity. SSIM evaluates the perceptual similarity between the reconstructed and original images, with higher scores indicating greater structural consistency. LPIPS, a deep learning-based metric, measures the perceptual similarity of the images, with lower values indicating higher perceptual quality. These metrics provide a comprehensive evaluation of \textit{SatelliteMaker}, ensuring both accuracy and perceptual quality in the reconstruction of remote sensing images.

\subsection{Comparison}
\textbf{Task-1: Comparison of Filling Methods}.
We compare multiple approaches for filling missing data in remote sensing images, including inpainting and diffusion-based methods. Palette \cite{saharia2022palette} and LaMa \cite{suvorov2022resolution} are inpainting methods designed for small missing regions, with LaMa performing better overall (see Table \ref{table_compare1}). However, these methods struggle with large missing areas.

Among diffusion models, Stable Diffusion (SD) \cite{rombach2022high} serves as the baseline model. While it achieves the highest SSIM (0.5402), the lack of topographic constraints results in unrealistic artifacts, leading to lower PSNR (17.1599) and higher RMSE (0.1448). ControlNet \cite{zhang2023adding} enhances SD by incorporating DEM as a conditional input, improving alignment with terrain features and achieving better PSNR (21.1847) and RMSE (0.0873).  

\textit{SatelliteMaker} further builds upon ControlNet by integrating the VGG-Adapter to enhance style consistency and reduce distribution discrepancies. Compared to ControlNet, \textit{SatelliteMaker} improves PSNR by {8.77\%} (from 21.1847 to 23.0438), reduces RMSE by {18.34\%} (from 0.0873 to 0.0713), and lowers MAE as well. Although its SSIM (0.4517) and LPIPS (0.3412) are slightly lower than those of ControlNet and LaMa, \textit{SatelliteMaker} excels in overall fidelity, ensuring better structural coherence and spatial consistency, making it the most robust solution.

\textbf{Task-2: Comparison of Temporal and Spatial Data Completion}.
We evaluate two scenarios: (1) temporal data completion using previous timestep inputs and (2) spatial data completion with DEM conditioning.

For temporal completion, interpolation methods perform poorly, with PSNR reaching only 11.9225. AutoEncoder \cite{rifai2011contractive} achieves the highest SSIM (0.6090) and PSNR (18.2038), making it suitable for small gaps but ineffective for large missing regions, as RMSE increases to 0.1230, degrading reconstruction quality. In DEM-based spatial completion, interpolation is not applicable, and diffusion models outperform traditional approaches. Among them, SD introduces artifacts, leading to lower SSIM (0.2819) and higher RMSE (0.1550), whereas ControlNet, benefiting from DEM conditioning, significantly improves PSNR (22.7866) and RMSE (0.0726).  

\textit{SatelliteMaker} surpasses ControlNet in all key metrics, increasing SSIM by {50.68\%} (from 0.3787 to 0.5704), improving PSNR by {6.83\%} (from 22.7866 to 24.3429), and reducing RMSE by {11.56\%} (from 0.0726 to 0.0642). These results highlight \textit{SatelliteMaker}'s superior capability in handling large-scale missing temporal and spatial data while maintaining structural and spectral consistency, leading to overall improved reconstruction quality.

\begin{table*}
    \centering
    \caption{Ablation study on the impact of VGG-Adapter module. \textbf{Bold} is the best result, and \underline{underline} represents the second-best result.}
    \resizebox{0.8\linewidth}{!}{
    \begin{tabular}{c|ccccc|ccccc}
    \toprule
        \multirow{2}{*}{\textbf{Band}} & \multicolumn{5}{c|}{\textbf{w/o VGG-Adapter}} & \multicolumn{5}{c}{\textbf{w/ VGG-Adapter}} \\ \cline{2-11}
        ~ & \textbf{SSIM$\uparrow$} & \textbf{PSNR$\uparrow$} & \textbf{RMSE$\downarrow$} & \textbf{MAE$\downarrow$} & \textbf{LIPIS$\downarrow$} & \textbf{SSIM$\uparrow$} & \textbf{PSNR$\uparrow$} & \textbf{RMSE$\downarrow$} & \textbf{MAE$\downarrow$} & \textbf{LIPIS$\downarrow$} \\ \midrule
        AVG & 0.3790  & 16.9564  & 0.1440  & 0.1216  & 0.1263  & 0.5704  & 24.3429  & 0.0642  & 0.0479  & 0.0469  \\ \hline
        Blue & \underline{0.3823}  & \textbf{17.9392}  & \textbf{0.1279}  & \textbf{0.1030}  & 0.1262  & 0.5685  & 23.9845  & 0.0657  & 0.0486  & \underline{0.0477}  \\ 
        Green & \textbf{0.3844}  & \underline{17.6690}  & \underline{0.1315}  & \underline{0.1068}  & \underline{0.1223}  & \underline{0.5783}  & \underline{24.0495}  & \underline{0.0653}  & \underline{0.0481}  & 0.0487  \\ 
        Red & 0.3674  & 15.2718  & 0.1743  & 0.1584  & 0.1377  & \textbf{0.5817}  & \textbf{26.2865}  & \textbf{0.0514}  & \textbf{0.0392}  & \textbf{0.0399}  \\ 
        NIR & 0.3816  & 16.9455  & 0.1424  & 0.1182  & \textbf{0.1189}  & 0.5530  & 23.0511  & 0.0746  & 0.0558  & 0.0513  \\ \bottomrule
    \end{tabular}
    }
    \label{table_ablation_vgg}
\end{table*}

\begin{table}[!ht]
    \centering
    \caption{Comparison of different methods in Task-1. \textbf{Bold} is the best result, and \underline{underline} represents the second-best result.}  
    \resizebox{1.0\linewidth}{!}{
    \begin{tabular}{cccccc}
    \toprule
        \textbf{Task-1} & \textbf{SSIM$\uparrow$} & \textbf{PSNR$\uparrow$} & \textbf{RMSE$\downarrow$} & \textbf{MAE$\downarrow$} & \textbf{LIPIS$\downarrow$} \\ \midrule
        SD & \textbf{0.5402}  & 17.1599  & 0.1448  & 0.0909  & 0.3392  \\ 
        Palette & 0.4333  & 18.6111  & 0.1212  & 0.0817  & 0.3442  \\ 
        LaMa & \underline{0.4937}  & 20.9301  & 0.0907  & 0.0637  & \underline{0.2902}  \\ 
        ControlNet & 0.4935  & \underline{21.1847}  & \underline{0.0873}  & \underline{0.0587}  & \textbf{0.2881}  \\ 
        \textit{SatelliteMaker} & 0.4517  & \textbf{23.0438}  & \textbf{0.0713}  & \textbf{0.0500}  & 0.3412  \\ \bottomrule
    \end{tabular}
    }
    \label{table_compare1}
\end{table}

\begin{table}[!ht]
    \centering
    \caption{Comparison of different methods in Task-2 with two input types: previous timestep data (upper part) and DEM (lower part). \textbf{Bold} is the best result, and \underline{underline} is the second-best result.}
    \resizebox{1.0\linewidth}{!}{
    \begin{tabular}{cccccc}
    \toprule
        \textbf{Task-2} & \textbf{SSIM$\uparrow$} & \textbf{PSNR$\uparrow$} & \textbf{RMSE$\downarrow$} & \textbf{MAE$\downarrow$} & \textbf{LIPIS$\downarrow$} \\ \midrule
        Interpolation & 0.5254  & 11.9225  & 0.2534  & 0.2023  & \underline{0.3380}  \\ 
        STCNN & \underline{0.4047}  & \underline{14.5317}  & \underline{0.1877}  & \underline{0.1498}  & 0.6835  \\ 
        Autoencoder & \textbf{0.6090}  & \textbf{18.2038}  & \textbf{0.1230}  & \textbf{0.0981}  & \textbf{0.2487}  \\ \hline
        Interpolation & - & - & - & - & - \\ 
        STCNN & 0.2232  & 13.9769  & 0.2107  & 0.1723  & 0.4208  \\ 
        Autoencoder & 0.2514  & 14.7913  & 0.2040  & 0.1684  & 0.3073  \\ 
        SD & 0.2819  & 16.1943  & 0.1550  & 0.1335  & 0.1468  \\ 
        ControlNet & \underline{0.3787}  & \underline{22.7866}  & \underline{0.0726}  & \underline{0.0570}  & \underline{0.0721}  \\ 
        \textit{SatelliteMaker} & \textbf{0.5704}  & \textbf{24.3429}  & \textbf{0.0642}  & \textbf{0.0479}  & \textbf{0.0469}  \\ \bottomrule
    \end{tabular}
    }
    \label{table_compare2}
\end{table}

\subsection{Ablation Study}
\textbf{VGG-Adapter Module.}  
The module corrects distribution discrepancies between generated and reference images, improving perceptual quality. Without the VGG-Adapter, noticeable artifacts such as brightness variations emerge in the generated images (see Figure \ref{fig_ablation_density}).

Quantitative analysis confirms the effectiveness of VGG-Adapter. For ControlNet, the brightness mean without the adapter is $\mu = 123.63$, significantly deviating from the ground truth ($\mu = 95.19$). With VGG-Adapter, SatelliteMaker aligns closely with the ground truth ($\mu = 95.27$), reducing distribution shifts.

Table \ref{table_ablation_vgg} further validates these improvements, showing consistent performance gains across all evaluation metrics. Notably, the red band, which previously exhibited discrepancies, saw the most substantial enhancement (see Figure \ref{fig_compare2}). These results highlight the VGG-Adapter’s ability to correct both perceptual and statistical inconsistencies, reinforcing its role as a critical component in \textit{SatelliteMaker}.

\textbf{Missing Ratio.}  
This ablation study evaluates model performance under varying missing data ratios, using PSNR and RMSE as key metrics. 
As shown in Figure \ref{fig_ablation_combined}, as the missing ratio increases from 10\% to 50\%, diffusion models remain relatively stable, while the performance of other models gradually declines.

Interpolation methods perform the worst, showing extreme sensitivity to missing data. PSNR drops sharply from 14.5317 (10\% missing) to around 5.0 (50\% missing), while RMSE rises from 0.1877 to approximately 0.9. This instability makes interpolation unsuitable for large-scale missing data scenarios. 

STCNN and AutoEncoder demonstrate moderate resilience, maintaining stable performance up to 30\% missing data. However, AutoEncoder's PSNR declines from 21.2371 at 10\% missing to 16.1943 at 50\% missing, indicating increasing degradation with higher missing ratios.

Diffusion models, particularly \textit{SatelliteMaker}, exhibit superior robustness across all missing ratios. \textit{SatelliteMaker} consistently outperforms other methods, achieving 24.7595 PSNR at 10\% missing and maintaining 23.7058 PSNR at 50\% missing, with the lowest RMSE throughout. Unlike statistical models that struggle with large gaps and deep learning models that suffer from stability and generalization issues, diffusion models—especially \textit{SatelliteMaker}—effectively mitigate these challenges. These results highlight \textit{SatelliteMaker} as the optimal solution for real-world remote sensing tasks where large missing data regions are prevalent.


\section{Conclusion}

We propose \textit{SatelliteMaker}, a diffusion-based model designed to address missing data in remote sensing images by filling spatial gaps within fixed time periods (Task-1) and temporal gaps at specific time points in spatiotemporal datasets (Task-2). Extensive experiments on Landsat 8 imagery from 10 countries and the EarthNet2021 dataset validate our approach, evaluated using SSIM, PSNR, RMSE, MAE, and LPIPS. The results demonstrate that \textit{SatelliteMaker} outperforms existing models, including SD, Palette, and LaMa in Task-1, and Interpolation, STCNN, AutoEncoder, and ControlNet in Task-2. \textit{SatelliteMaker} effectively fills spatial and temporal gaps in remote sensing images and demonstrates {stable performance}. The model incorporates {DEM as a condition input}, leveraging its strong correlation with surface cover to enhance content generation and providing a {robust solution} for large-scale missing data scenarios. Additionally, we introduce a {VGG-Adapter module} based on distribution loss, which {minimizes distributional discrepancies} and ensures {style consistency}, achieving {state-of-the-art} performance across tasks. 
\textit{SatelliteMaker} offers a {scalable solution} to remote sensing data challenges.

{
    \small
    \bibliographystyle{ieeenat_fullname}
    \bibliography{output}
}


\end{document}